\documentclass{article}

\usepackage{arxiv}

\usepackage[utf8]{inputenc} 
\usepackage[T1]{fontenc}    
\usepackage{hyperref}       
\usepackage{url}            
\usepackage{booktabs}       
\usepackage{amsfonts}       
\usepackage{nicefrac}       
\usepackage{microtype}      
\usepackage{lipsum}
\usepackage{graphicx}
\usepackage{amsmath, natbib, cases}
\usepackage{multirow, boldline}

\graphicspath{ {./images/} }

\title{Investigation of the Sense of Agency in Social Cognition, based on frameworks of Predictive Coding and Active Inference: A simulation study on multimodal imitative interaction}

\author{
  Wataru Ohata and Jun Tani\thanks{Corresponding author}\\
  Cognitive Neurorobotics Research Unit\\
  Okinawa Institute of Science and Technology Graduate University, Okinawa, Japan\\
  \texttt{\{wataru.ohata, jun.tani\}@oist.jp}
}

\begin{document}
\maketitle
\begin{abstract}
When agents interact socially with different intentions (or wills), conflicts are difficult to avoid.   
Although the means by which social agents can resolve such problems autonomously has not been determined, dynamic characteristics of agency may shed light on underlying mechanisms. Therefore, the current study focused on the sense of agency, a specific aspect of agency referring to congruence between the agent's intention in acting and the outcome, especially in social interaction contexts. Employing predictive coding and active inference as theoretical frameworks of perception and action generation, we hypothesize that regulation of complexity in the evidence lower bound of an agent's model should affect the strength of the agent's sense of agency and should have a significant impact on social interactions. To evaluate this hypothesis, we built a computational model of imitative interaction between a robot and a human via visuo-proprioceptive sensation with a variational Bayes recurrent neural network, and simulated the model in the form of pseudo-imitative interaction using recorded human body movement data, which serve as the counterpart in the interactions. A key feature of the model is that the complexity of each modality can be regulated differently by changing the values of a hyperparameter assigned to each local module of the model. We first searched for an optimal setting of hyperparameters that endow the model with appropriate coordination of multimodal sensation. These searches revealed that complexity of the vision module should be more tightly regulated than that of the proprioception module because of greater uncertainty in visual information flow. Using this optimally trained model as a default model, we investigated how changing the tightness of complexity regulation in the entire network after training affects the strength of the sense of agency during imitative interactions. The results showed that with looser regulation of complexity, an agent tends to act more egocentrically, without adapting to the other. In contrast, with tighter regulation, the agent tends to follow the other by adjusting its intention. 
We conclude that the tightness of complexity regulation significantly affects the strength of the sense of agency and the dynamics of interactions between agents in social settings. 


\end{abstract}

\section{Introduction}
Humans are social beings by nature, and each individual regularly interacts with others in various ways. Even though individuals act based on their intentions or wills, they sometimes acts in agreement with others, doing something collaboratively, while at other times they disagree. Either case may be conscious or unconscious. What determines such the type of interaction and how? To evaluate this problem, we consider possible relationships between {\it agency} of each individual and social interactions between individuals. Then, we introduce predictive coding and active inference to formulate the problem in a computational framework and we propose a specific hypothesis to predict the type of interaction. We deliver a schematic of our computational model and experimental setup to evaluate the hypothesis and conclude the section by highlighting some critical findings.

\medskip
\subsection{Agency in social cognition}
In social interactions, agents sometimes cooperate by sharing intentions so as to derive mutual benefits, while at other times they cause conflicts by following their own intentions and ignoring the interests of others. Although how such complexities in social interactions emerge is not obvious, we hypothesized that dynamic characteristics of {\it agency} in social interactions might shed light on underlying mechanisms. Recently, the study of agency has attracted considerable attention from researchers in various disciplines, including philosophy, psychology, cognitive science, and neuroscience. Specifically, the sense of agency (SoA) \citep{gallagher2000philosophical, synofzik2008beyond, moore2009modulating} refers to congruence between an agent's intention or belief in an action and its anticipated outcome, which endows the agent with the sense that {\it ``I am the one generating this action''}. Along with studies in experimental psychology, building a computational model of SoA is also important in order to explore the nature of agency \citep{legaspi2019bayesian}. In the study of computational models of agents, predictive coding (PC) \citep{rao1999, tani1999learning, lee2003, friston2005theory, hohwy2013predictive, clark2015, friston2018} and active inference (AIF) \citep{friston2009, friston2010action, baltieri2017, buckley2017, pezzulo2018hierarchical, oliver2019active} have recently attracted considerable attention since they provide rigid theoretical frameworks for defining perception and action generation. In the framework of PC and AIF, an agent's intention or belief can be formulated as a predictive model, and it is thought that congruence between the prediction of action outcomes and observations reinforces the SoA \citep{friston2012prediction}.

In situations involving social interaction, however, where multiple agents interact, it becomes challenging for each agent to sustain its SoA, because other agents, having their own intentions, may not act as desired. If social agents are required to coordinate actions so as to obtain benefits by minimizing possible conflicts, we speculated that the strength of agency should be arbitrated among those agents during some conflicts. Let us consider a dyadic synchronized imitation as an example of social interaction, wherein two agents attempt to synchronously imitate one another's movement patterns using predictions based on prior learning. In addition, let us assume a setting in which two agents imitate one another in sequences of movement patterns based on memorized transition rules, in which unpredictable transitions in movement patterns are included. For example, either movement pattern B or C can appear after movement pattern A (see also Figure \ref{data preparation} (A)). In this setting, agent 1 may opt for movement pattern B after A, acting as a leader with {\it strong agency} and agent 2 may just follow agent 1 by generating pattern B with {\it weak agency}. This can result in successful mutual imitation without generating conflict. However, if both agents maintain strong agency, each may generate its own pattern (B or C) without compromise, resulting in conflict.

While investigating agency in social interactions, we concluded that it would also be worthwhile to consider how agency and mirror neuron systems (MNS) \citep{rizzolatti2014mirror, kilner2007predictive} might be related, since MNS are thought to contribute to various types of social cognitive behavior, including imitation \citep{hurley2005perspectives}. MNS was first discovered in area F5 of the monkey premotor cortex \citep{di1992understanding, gallese1996action}, and it is activated when monkeys execute their own actions, as well as when observing those performed by others. Because MNS uses observations of an action to generate the same action, it may participate in imitative behaviors, which are thought to be the basis of various higher cognitive functions \citep{aly2015online, kohler2002hearing, oztop2006mirror, oztop2013mirror}. A natural question regarding such an MNS mechanism is how the agency of each individual can be exerted if MNS is the default mode. Intention to generate an action could conflict with an automatic response to imitate an action demonstrated by others.
Although modeling studies of MNS have also been conducted from the view point of PC and AIF using Bayesian frameworks \citep{friston2011action, kilner2007predictive} and by using deterministic recurrent neural networks (RNNs) \citep{ito2004line, ahmadi2017can, hwang2020dealing}, the aforementioned problem of agency has not been well considered.

\medskip
\subsection{Predictive coding and active inference}
Next, let us consider how the strength of agency can be modeled using a framework of PC and AIF. For this purpose, first we briefly review the concepts of PC and their mathematical properties, as follows. In PC, perception is thought to be achieved via iterative interactions between a prior expectation of a sensation and a posterior inference from a sensory outcome. The prior expectation of the sensation can be modeled by statistical generative models that map the prior of the latent variable to the sensory expectation. The posterior inference can be carried out by taking the error between the expected sensation and its outcome and by updating the posterior of the latent variable in the direction of minimizing the error, under the constraint of minimizing Kullback-Leibler divergence (KL divergence) between the posterior distribution and that of the prior. Typically, both the prior and the posterior are represented by Gaussian distributions with parameters of mean and variance, as will be described later. This is equal to maximizing the lower bound of the logarithm of marginal likelihood (a.k.a evidence lower bound) expressed by two terms: {\it accuracy} and {\it complexity}.
\postdisplaypenalty=0
\footnotesize
\begin{align}
    \ln p_\theta(\boldsymbol{X})&\geq\underbrace{\int q_\phi(\boldsymbol{z}|\boldsymbol{X})\ln\frac{p_\theta(\boldsymbol{X},\boldsymbol{z})}{q_\phi(\boldsymbol{z}|\boldsymbol{X})}d\boldsymbol{z}}_{\rm Evidence\ lower\ bound}\\
    &=\underbrace{\mathbb{E}_{q_\phi(\boldsymbol{z|\boldsymbol{X}})}[\ln p_\theta(\boldsymbol{X}|\boldsymbol{z})]}_{\rm Accuracy}-\underbrace{D_{\rm KL}[q_\phi(\boldsymbol{z}|\boldsymbol{X})\Vert p(\boldsymbol{z})]}_{\rm Complexity}\label{elbo}
\end{align}
\normalsize
where $\boldsymbol{X}$ is the observation, $\boldsymbol{z}$ is the latent variable, $q_\phi(\boldsymbol{z}|\boldsymbol{X})$ is the approximate posterior, and $\theta$ and $\phi$ are the parameters of the model. {\it Accuracy} is the expectation of log-likelihood with respect to the approximate posterior, which represents reconstruction of the observation with the approximate posterior. {\it Complexity} is the KL divergence between the approximate posterior and the prior, which serves to regularize the model. Importantly, in maximizing the lower bound, the interplay between these two terms characterizes how the model behaves in learning and prediction \citep{higgins2017beta}. Maximization of the lower bound is equivalent to minimization of {\it free-energy} proposed by Friston \citep{friston2005theory}.

Next, AIF is described briefly. AIF explains that action or motor commands should be generated so that their sensory outcomes coincide with expected outcomes. As a simple example, consider how expected proprioception in terms of robot joint angles can be achieved by generating sufficient motor torque. This can be done with an inverse model that maps expected joint angles to the required motor torques, or by employing a PID controller such that necessary motor torque to minimize errors between expected joint angles and actual angles can be derived by means of a simple error feedback mechanism. Both PC and AIF attempt to minimize error between the expected sensation and the actual outcome; however, in PC this is accomplished by changing the intention via the posterior inference and by changing the environment state through action in AIF. When PC and AIF are performed in tandem, while an agent acts on the environment, an agent with a more precise prior (smaller variance) should behave with strong agency, being less likely to change its own intention, and more likely to change the environmental state. On the other hand, an agent with a less precise prior (with larger variance) should behave with weaker agency, being more likely to change its own intention than the environmental state.

\medskip
\subsection{Related work}
Although PC and AIF have attracted much attention from brain modeling researchers, it is unusual to see them used in computational studies employing learnable neural network models, especially those that can handle continuous spatio-temporal patterns characterized in multimodal sensory inputs. To this intent, Ahmadi and Tani \citep{ahmadi2019novel} recently proposed so-called, Predictive-coding-inspired Variational Recurrent Neural Network (PV-RNN). PV-RNN is a type of variational recurrent neural network that approximates the posterior at each time step in sequential patterns with variational inference, and is formalized by employing predictive coding. By making predictions in the form of the sequential prior \citep{chung2015recurrent} with time-varying parameterized Gaussian distribution, PV-RNN enables the model to represent strength of intention or agency. \cite{ahmadi2019novel} introduced a hyper parameter $w$ called the meta-prior, which weights regulation of the complexity term in the evidence lower bound (the second term in equation \ref{elbo}). They found that a model trained with looser regulation of the complexity term, achieved by setting the meta-prior to a larger value, develops more deterministic dynamics with higher estimated precision in the sequential prior, whereas a model trained with tighter regulation, accomplished by setting the meta-prior to a smaller value, develops more probabilistic dynamics with lower estimated precision. In another attempt to implement free-energy minimization with an artificial neural network, Pitti et al. \citep{pitti2020gated} proposed a spiking neural network architecture that minimizes free-energy to model the fronto-striatal system in the brain.

Chame and Tani \citep{chame2019cognitive} used PV-RNN to conduct a human-robot interaction experiment using a single perceptual channel of proprioception. Although their analysis of the experiments was preliminary, they suggested that when the model is trained under looser regulation of the complexity term, the model tends to behave egocentrically, adapting less to proprioceptive inputs, whereas under tighter regulation of the complexity term, the network tends to behave more passively, adapting more to proprioceptive inputs. However, such network characteristics, once developed through learning under particular conditions to regulate the complexity term, cannot be changed thereafter. 
In social interactions, it is natural that agents act differently, depending on the social context at a given moment. Sometimes they tend to preserve their prior intention by acting perversely, and at other times they change it more easily by adapting to intentions of others. 
The current study examines whether such shifts in strength of agency can be achieved during the interaction phase by changing the value of the meta-prior from the default strength set in the learning phase.

\medskip
\subsection{Imitative interaction using a variational Bayes recurrent neural network}
Here, we explain the general concept underlying our computational model, experimental design, and obtained results. We first proposed an artificial neural network model that can be applied to an imitative interaction task using multimodal sensation of vision and proprioception by extending PV-RNN. PV-RNN is used because to our knowledge this network model is the only RNN-type model that can instantiate predictive coding and active inference in a continuous spatio-temporal domain by following a Bayesian framework. The proposed model is comprised of a multi-layered PV-RNN with a branching structure, in which two branches responsible for perception of vision and proprioception are connected through an associative module. In addition, the current model inherits the structure of Multiple Timescale Recurrent Neural Network (MTRNN) \citep{yamashita2008}. MTRNN extracts a temporal hierarchy contained in sequential patterns \citep{yamashita2008, nishimoto2009development, hwang2020dealing}. By assigning faster timescales to the peripheral sensory modules for vision and proprioception and slower timescales to the associative module, hierarchical multimodal integration from sensory-motor levels to abstract intention levels should be achieved.

The entire network model is considered a generative model that predicts incoming visual sensation and proprioception simultaneously through a generative process along with a top-down pathway from the associative module to both of the sensory modules. The resultant prediction error for each sensory modality is back-propagated through time \citep{werbos1974beyond, rumelhart1985learning} (BPTT) and through each module to the associative module, by which the latent state in each module is modulated so as to maximize the evidence lower bound shown in the equation \ref{elbo}. This corresponds to the posterior inference. The network is trained through supervised learning by maximizing the evidence lower bound. 

However, coordinating multimodal sensations appropriately is still not an easy problem when intrinsic complexity and randomness in spatio-temporal patterns differ in each modality \citep{ogata2010inter, valentin:hal-02295330}. Studies on cue integration in multimodal sensation have shown that inferences about the hidden state of the environment should be accomplished by assigning the greatest weight to information obtained from the most reliable sensory modality \citep{battaglia2003bayesian}. In predictive coding, reliability can be represented by the accuracy estimated for each modality of the sensory model, provided that its generalization is preserved by minimizing model complexity adequately when the amount of training data is limited. We speculate that the complexity term should be regulated adequately for each sensory modality during training, such that the best generalization can be achieved for each. Since each PV-RNN module can be assigned different values of the meta-prior, the above could be achieved by searching for an adequate value of each meta-prior though trial and error during the learning phase.

The proposed model was evaluated by simulating ``pseudo" imitative interaction using visuo-proprioceptive sequence patterns recorded from human demonstrators. Although human-robot interaction should be studied in a physical system to allow the human and the robot to respond to each other in an online fashion, it is difficult for the current system to work in real time because inference of the posterior using PV-RNN is computationally intensive, especially when pixel-level vision is used as one of the sensory modalities. Therefore, the current study focuses on simulation experiments using pre-recorded data.

First, we investigated how changing the tightness used to regulate the complexity term for each sensory module in the learning phase affects the quality of integrating multimodal sensation in an imitative interaction. For this purpose, we examined possible effects of assigning different values of the meta-prior to the vision module and the proprioceptive module, on performance characteristics in learning, as well as in the resulting imitative interaction. Our results suggest that regulating complexity more in the vision module than in the proprioception module facilitates better imitation performance in multi-modal sensation after learning, because visual sensory information contains more randomness than proprioceptive information. 

Second, as the main motivation of the current study, we investigated how changing the tightness used to regulate the complexity term in the entire network after the learning phase affects the strength of agency. Using a network trained by tuning the meta-priors assigned to each sensory module in the previous experiment, we examined how increasing or decreasing meta-prior values throughout the network compared to those used during learning affects imitative behavior. We found that a network that tightly regulates the complexity term by setting smaller values of the meta-prior tends to follow human movement patterns by adapting its internal states. On the other hand, the network that loosely regulates the complexity term by setting larger values of the meta-prior tends to generate more egocentric/self-centered movement patterns with less sensitivity to changes or fluctuations in human movement patterns by adapting its internal state less. The current paper presents a detailed analysis of the underlying mechanisms accounting for these observed phenomena. 

Below, the {\it Model} section details the proposed model. It describes an overall system, learning process, derivation of the evidence lower bound of the proposed model, how the trained model was tested in pseudo imitative interaction, and implementation of the model. The {\it Experiment} section explains the experimental design, procedures of data preparation, and the results of the two experiments. The {\it Discussion} section summarizes the experimental results and discusses their implications.

\section{Model}

\subsection{Model overview}
\label{model overview}
This subsection describes briefly how multimodal imitative interaction of agents perceiving visuo-proprioceptive sensory inputs can be modeled using concepts of predictive coding and active inference. Among various types of imitation, synchronized imitation is considered in the current study by virtue of its simplicity. In synchronized imitation, the agent is required to imitate target patterns demonstrated by its counterpart by predicting them on the basis of prior learning. Although target patterns to imitate are structurally the same as previously learned patterns, they could involve marginal variations, as in speed, amplitude, and shape. Synchronized imitation can be achieved by means of iterative cycling of sensory input predictions during the demonstration, generation of corresponding movement, and updating the latent state of the network using the resulting sensory prediction error. To generate movement, one step, look-ahead prediction of proprioception is fed into an inverse model \citep{kawato1987hierarchical}, which is often implemented by a PID feedback controller in robots. A PID feedback controller computes an optimal motor torque as the motor command to minimize the error between the predicted proprioception (the target joint angles) and the actual proprioception (the actual joint angles). This corresponds to active inference \citep{friston2010action, friston2011action}, as described previously. The latent state can be updated using a scheme called error regression \citep{tani1999learning, ito2004line, hwang2020dealing, ahmadi2019novel}, by which sensory perception assumed in a predictive coding framework can be performed.

\begin{figure}[ht]
    \centering
    \includegraphics[width=1.0\textwidth]{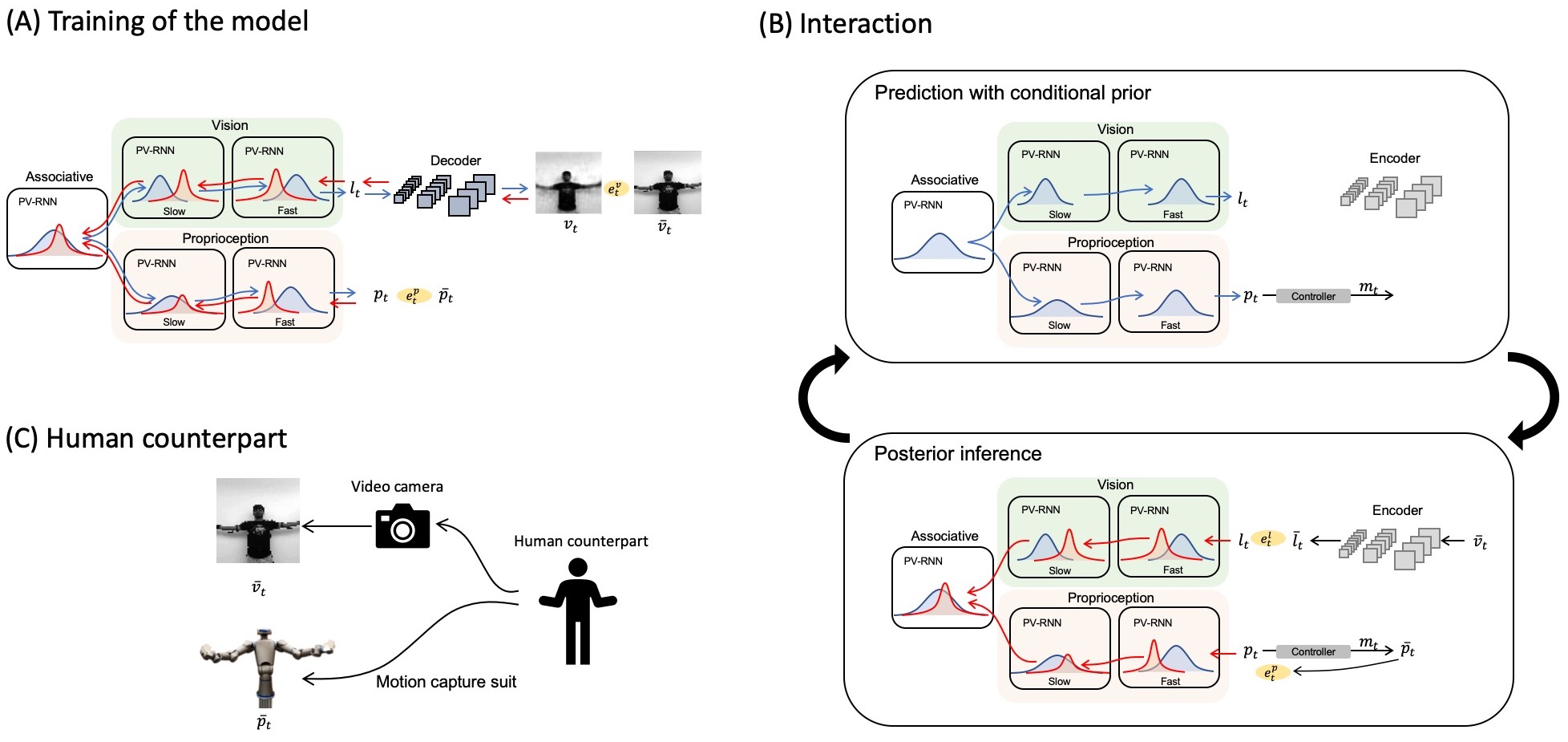}
    \caption{Overall schematic of the proposed model. Blue and red bell curves represent prior and posterior distributions, respectively. Blue and red arrows illustrate information flows of the prediction with conditional prior and posterior inferences, respectively. (A) The training scheme of the proposed network model. (B) The cycle of prediction with conditional prior and posterior inferences during an interaction with a human. (C) A diagram of providing the configuration of a human counterpart to the network.}
    \label{model}
\end{figure}

Now we look at how the PV-RNN \citep{ahmadi2019novel} can be used to implement the model for multimodal imitative interaction of a robot agent receiving visuo-proprioceptive sensory inputs based on frameworks of predictive coding and active inference. Figure \ref{model} shows the overall system view, consisting of a PV-RNN, a robot, and a human counterpart. The human demonstrates movement patterns to the robot both visually and kinesthetically, guiding the robot's posture via a motion capture suit. Unfortunately, it was infeasible for the proposed system to work stably in real-time because posterior inference using an error-regression scheme, detailed in section \ref{error-regression}, requires intensive computation. Hence, in the current study, we simulated the imitative interaction between a human and a robot shown in Figure \ref{model} as a pseudo imitative interaction in which pre-recorded body movements sampled from a human serve as the robot counterpart using the setting shown in Figure \ref{model} (C).

PV-RNN is considered a generative model, formulated in a continuous spatio-temporal domain, employing a variational Bayes framework, as described previously. It infers the posterior at each time step using variational inference, in which the reconstruction error is minimized with regularization of the KL divergence between the inferred posterior and the conditional prior. This is implemented by means of a so-called {\it error-regression scheme}, detailed in section \ref{error-regression}.

A PV-RNN inherits the concept of a Multiple Timescale Recurrent Neural Network (MTRNN)\citep{yamashita2008}, which is characterized by its architecture because it allocates different timescale dynamics to different layers. Higher layers are endowed with slower timescale dynamics and lower layers with faster dynamics, as inspired by recent cognitive neuroscience evidence \citep{newell2001time,huys2004multiple,smith2006interacting,kording2007dynamics}. Introduction of multiple timescale dynamics can enhance abstraction and generalization in learning by extracting action-primitive hierarchies or chunking structures from observed multimodal sensory inputs \citep{yamashita2008, choi2018predictive, hwang2020dealing}.

These characteristics of variational Bayes frameworks and MTRNN enable PV-RNN to utilize hierarchically organized probabilistic representation, i.e., while the network extracts a hierarchical structure from an observation, it also assigns a different degree of uncertainty within the hierarchy. For example, given a task in which the network is required to predict a sequence of body movements comprised of a small number of primitive patterns, the network can be certain about details of the primitive patterns, but less certain about the sequence of the primitives. In such a case, the lower level of the network responsible for prediction of details of each primitive movement shows small uncertainty, while the higher level in charge of prediction of the order of those primitive patterns shows high uncertainty.

Sensory modules for proprioception and vision were modeled with multi-layered PV-RNNs and modules were connected via an associative module, also based on a PV-RNN. Figure \ref{model} (A) depicts a schematic of the proposed model and how it is trained. 
The associative module generates the prior, conditioned by the latent state at the previous time step in this module. The prior is then fed to both the proprioception and vision modules along the top-down pathway. Each sensory module also generates a prior at each time step conditioned by the previous latent state of the module, computed using top-down information provided by the associative module, by which predictions of sensory inputs, proprioception and vision, are generated in the subsequent time step. Note that the vision module predicts a low-dimensional vector, which is then fed to a CNN-type decoder \citep{lecun1989backpropagation, lecun1998gradient} to generate actual pixel visual images, in order to reduce computational costs.

A dataset of visuo-proprioceptive patterns demonstrated by human participants is used to train the model. To generate these data, a human wearing a motion-capture suit demonstrates body movements while simultaneously recording a video. The motion capture suit maps the human's body configuration into the humanoid robot's joint angle values. These synchronized joint-angle trajectories and videos serve as the target of the model. The whole network is optimized simultaneously so as to maximize the evidence lower bound of the model via BPTT. The design of body movement patterns used in this study is detailed in section \ref{data preparation}.

Figure \ref{model} (B) describes how the trained model performs imitative interactions. An imitative interaction involves a cycle of predictions with conditional prior and posterior inferences. At each time step, the network predicts proprioception $\boldsymbol{p}_t$ and a low-dimensional latent representation of vision $\boldsymbol{l}_t$ with the prior conditioned by the latent variable in each module at the previous time step. The proprioceptive prediction $\boldsymbol{p}_t$ is supplied to the controller, followed by computation of motor commands $\boldsymbol{m}_t$ to achieve the expected joint positions and generation of the movement. Then, a new visual image and proprioception are acquired. The raw pixel image is fed to a CNN-type encoder that has been separately trained to obtain the target for the low-dimensional latent representation $\bar{\boldsymbol{l}_t}$. Resultant prediction errors $e^l_t$ and $e^p_t$ are computed in vision and proprioception, respectively, which are then used to infer the posterior in each PV-RNN layer with regulation of the KL divergence between the inferred posterior and the conditional prior so that the lower bound is maximized by BPTT. This optimization process to infer the posterior is iterated a fixed number of times at each sensory-motor sampling time step, and the optimized posterior is used to make the best prediction with the conditional prior to the succeeding time step.

Figure \ref{model} (C) denotes how the robot's network model senses movement patterns demonstrated by the human counterpart. 

\medskip
\subsection{Derivation of evidence lower bound}
PV-RNN is a generative, inference model based on the graphical representation shown in Figure \ref{er} (this figure will be explained in detail in section \ref{error-regression}). It is comprised of deterministic variables $\boldsymbol{d}$, i.e., assumed to follow Dirac delta distributions, and stochastic variables $\boldsymbol{z}$. The model includes a prior and infers the corresponding posterior by variational inference. We modified the original PV-RNN at four points with respect to dependencies of variables. First, in our model, there are no connections between the output of the network $\boldsymbol{x}$ and $\boldsymbol{z}$, which exist in the original PV-RNN. This is for simplification of the model, and it was confirmed that removing these connections did not hinder network performance. Second, the current network does not have connections from the lower layer to the higher layer, which the original network does have. This modification is intended to separate more clearly the information flow between top-down generative prediction and bottom-up error propagation. Third, diagonal connections from the higher layer during the previous time step to the lower layer during the succeeding time step are changed to vertical connections during the same time step. Last, the prior distribution of $\boldsymbol{z}_t$ at time step 1 has been changed. In the original study, the distribution is simply mapped from $\boldsymbol{d}_0$. In the current study, however, it is assumed that $p(\boldsymbol{z}_1)$ follows a unit Gaussian distribution to control the initial sensitivity of the model. Following derivation of the evidence lower bound in \cite{ahmadi2019novel} and considering the introduction of the unit Gaussian at time step $1$, the evidence lower bound of the proposed visuo-proprioceptive model is derived as
\footnotesize
\begin{equation}
\begin{split}
    \ln(\boldsymbol{p}_{1:T}, \boldsymbol{v}_{1:T}|\boldsymbol{d}_0^{*})&\geq\sum_{t=1}^T\biggl\{\mathbb{E}_{q^a, q^p}[\ln P(\boldsymbol{p}_t|\boldsymbol{d}_t^{p, 1})]+\mathbb{E}_{q^a, q^v}[\ln P(\boldsymbol{v}_t|\boldsymbol{d}_t^{v, 1})]\biggl\}\\
    &\quad -\sum_{l\in A}D_{\rm KL}[q(\boldsymbol{z}_1^l|\boldsymbol{d}_0^l, e^p_{1:T}, e^v_{1:T})\Vert p(\boldsymbol{z}^u)]-\sum_{l\in P}D_{\rm KL}[q(\boldsymbol{z}_1^l|\boldsymbol{d}_0^l, e^p_{1:T})\Vert p(\boldsymbol{z}^u)]\\
    &\quad -\sum_{l\in V}D_{\rm KL}[q(\boldsymbol{z}_1^l|\boldsymbol{d}_0^l, e^v_{1:T})\Vert p(\boldsymbol{z}^u)]+\sum_{t=2}^T\biggl\{-\sum_{l\in A}D_{\rm KL}[q(\boldsymbol{z}_t^{l}|\boldsymbol{d}_{t-1}^{l}, e_{t:T}^p, e_{t:T}^v)\Vert p(\boldsymbol{z}_t^{l}|\boldsymbol{d}_{t-1}^{l})]\\
    &\quad\quad-\sum_{l\in P}D_{\rm KL}[q(\boldsymbol{z}_t^{l}|\boldsymbol{d}_{t-1}^{l}, e_{t:T}^p)\Vert p(\boldsymbol{z}_t^{l}|\boldsymbol{d}_{t-1}^{l})]-\sum_{l\in V}D_{\rm KL}[q(\boldsymbol{z}_t^{l}|\boldsymbol{d}_{t-1}^{l}, e_{t:T}^{v})\Vert p(\boldsymbol{z}_t^{l}|\boldsymbol{d}_{t-1}^{l})]\biggl\}
\end{split}
\end{equation}
\normalsize
where A, P, and V represent the associative module, the proprioception module, and the vision module, respectively, and $l$ indicates the index of a layer in each module. $\boldsymbol{p}_{1:T}$ and $\boldsymbol{v}_{1:T}$ are time series propriocetive and visual patterns. $\boldsymbol{d}_0^*$ represents $\boldsymbol{d}$ in all layers at time step 0. $\mathbb{E}_{q^a, q^p}$ denotes the expectation over all distributions of $\boldsymbol{z}$ in the associative module and the proprioception module, and $\mathbb{E}_{q^a, q^v}$ denotes expectation over all distributions of $\boldsymbol{z}$ in the associative module and the vision module. $\boldsymbol{d}_t^{p,1}$ is the deterministic variable in the lowest layer of the proprioception module at time step $t$, and $\boldsymbol{d}_t^{v, 1}$ is that in the lowest layer of the vision module. $\boldsymbol{z}_t^l$ is the stochastic variable at time step $t$ in the $l$th layer in each module. $e^p_{t:T}$ and $e^v_{t:T}$ are the prediction errors between the predicted patterns and the target patterns at time step from $t$ to $T$ in proprioception and vision, respectively. $p(\boldsymbol{z}^u)$ indicates the unit Gaussian distribution serving as the prior at time step $1$. By introducing the meta-prior, which weights the KL divergence between the approximate posterior and the prior in a layer-specific manner, the evidence lower bound of the model is defined as
\footnotesize
\begin{equation}
\begin{split}
    \mathcal{L}_w&:=\sum_{t=1}^T\biggl\{\underbrace{\mathbb{E}_{q^a, q^p}[\ln P(\boldsymbol{p}_t|\boldsymbol{d}_t^{p, 1})]}_{\rm Accuracy\ in\ proprioception}
    +\underbrace{\mathbb{E}_{q^a, q^v}[\ln P(\boldsymbol{v}_t|\boldsymbol{d}_t^{v, 1})]}_{\rm Accuracy\ in\ vision}\biggl\}\\
    &\quad -\sum_{l\in A}w^l_1\underbrace{D_{\rm KL}[q(\boldsymbol{z}_1^l|\boldsymbol{d}_0^l, e^p_{1:T}, e^v_{1:T})\Vert p(\boldsymbol{z}^u)]}_{{\rm Complexity\ in\ associative\ module}}
    -\sum_{l\in P}w^l_1\underbrace{D_{\rm KL}[q(\boldsymbol{z}_1^l|\boldsymbol{d}_0^l, e^p_{1:T})\Vert p(\boldsymbol{z}^u)]}_{{\rm Complexity\ in\ proprioception\ module}}\\
    &\quad-\sum_{l\in V}w^l_1\underbrace{D_{\rm KL}[q(\boldsymbol{z}_1^l|\boldsymbol{d}_0^l, e^v_{1:T})\Vert p(\boldsymbol{z}^u)]}_{{\rm Complexity\ in\ vision\ module}}
    +\sum_{t=2}^T\biggl\{-\sum_{l\in A}w^{l}\underbrace{D_{\rm KL}[q(\boldsymbol{z}_t^{l}|\boldsymbol{d}_{t-1}^{l}, e_{t:T}^p, e_{t:T}^v)\Vert p(\boldsymbol{z}_t^{l}|\boldsymbol{d}_{t-1}^{l}))]}_{\rm Complexity\ in\ associative\ module}\\
    &\quad\quad-\sum_{l\in P}w^{l}\underbrace{D_{\rm KL}[q(\boldsymbol{z}_t^{l}|\boldsymbol{d}_{t-1}^{l}, e_{t:T}^p)\Vert p(\boldsymbol{z}_t^{l}|\boldsymbol{d}_{t-1}^{l})]}_{\rm Complexity\ in\ proprioception\ module}
    -\sum_{l\in V}w^{l}\underbrace{D_{\rm KL}[q(\boldsymbol{z}_t^{l}|\boldsymbol{d}_{t-1}^{l}, e_{t:T}^{v})\Vert p(\boldsymbol{z}_t^{l}|\boldsymbol{d}_{t-1}^{l})]}_{\rm Complexity\ in\ vision\ module}\biggl\}
\end{split}
\end{equation}
\normalsize
where $w_1^l$ indicates the meta-prior in the $l$th layer at $t=1$ in the associative module, the proprioception module, and the vision module, respectively. $w^{l}$ represents the meta-priors in the $l$th layer after $t=2$ in each module. Parameters of the model are optimized by maximizing the lower bound, which corresponds to minimizing the free energy.

\medskip
\subsection{Learning process}
It is noted that unlike other models employing online learning methods \citep{boucenna2014robot,boucenna2016robots}, our model is trained offline with pre-recorded dataset. The entire network model is trained by maximizing the evidence lower bound. Thus, given the time step length $T$ of proprioceptive patterns $\boldsymbol{p}_{1:T}$ and visual patterns $\boldsymbol{v}_{1:T}$, the cost function to be minimized is defined as
\footnotesize
\begin{equation}
    \begin{split}
        {\rm cost}:=&\sum_{t=1}^{T}\biggl\{\frac{1}{2R^p}\left\Vert \boldsymbol{p}_t-\bar{\boldsymbol{p}_t}\right\Vert^2
        +\frac{1}{2R^v}\left\Vert \boldsymbol{v}_t-\boldsymbol{\bar{v}}_t\right\Vert^2\\
        &\quad +\sum_{l\in A}\frac{w^l_1}{R^l}D_{\rm KL}[q(\boldsymbol{z}_1^l|\boldsymbol{d}_0^l, e^p_{1:T}, e^v_{1:T})\Vert p(\boldsymbol{z}^u)]
        +\sum_{l\in P}\frac{w^l_1}{R^l}D_{\rm KL}[q(\boldsymbol{z}_1^l|\boldsymbol{d}_0^l, e^p_{1:T})\Vert p(\boldsymbol{z}^u)]\\
        &\quad +\sum_{l\in V}\frac{w_1^l}{R^l}D_{\rm KL}[q(\boldsymbol{z}_1^l|\boldsymbol{d}_0^l, e^v_{1:T})\Vert p(\boldsymbol{z}^u)]|
        +\sum_{t=2}^T\biggl\{\sum_{l\in A}\frac{w^l}{R^{l}}D_{\rm KL}[q(\boldsymbol{z}_t^{l}|\boldsymbol{d}_{t-1}^{l}, e_{t:T}^p, e_{t:T}^v)\Vert p(\boldsymbol{z}_t^{l}|\boldsymbol{d}_{t-1}^{l})]\\
        &\quad +\sum_{l\in P}\frac{w^l}{R^{l}}D_{\rm KL}[q(\boldsymbol{z}_t^{l}|\boldsymbol{d}_{t-1}^{l}, e_{t:T}^p)\Vert p(\boldsymbol{z}_t^{l}|\boldsymbol{d}_{t-1}^{l})]
        +\sum_{l\in V}\frac{w^l}{R^{l}}D_{\rm KL}[q(\boldsymbol{z}_t^{l}|\boldsymbol{d}_{t-1}^{l}, e_{t:T}^{v})\Vert p(\boldsymbol{z}_t^{l}|\boldsymbol{d}_{t-1}^{l})]\biggl\}
    \end{split}
\end{equation}
\normalsize
where A, P, and V represent the associative module, the proprioception module, and the vision module. $R^p$ and $R^v$ are the dimensions of proprioceptive patterns and visual patterns to normalize prediction errors, and $R^{l}$ is the dimension of the distributions of $\boldsymbol{z}$ to normalize the KL divergence. Each output in the vision and proprioceptive modules is represented by a multivariate Gaussian distribution with an estimation of the mean for each dimension and covariant matrix as the identity matrix, for simplicity. This leads to minimization of the mean squared error, which is an estimator of the log-likelihood in the accuracy term when maximizing the lower bound. 

Since the prior and posterior distributions are assumed to follow a multivariate Gaussian distribution with a diagonal covariant matrix, the KL divergence in the cost function is analytically computed. Given two $n$ dimensional multivariate Gaussian distributions $p(\boldsymbol{z})=\mathcal{N}(\boldsymbol{z};\boldsymbol{\mu}^p,\boldsymbol{\sigma}^p)$ and $q(\boldsymbol{z})=\mathcal{N}(\boldsymbol{z};\boldsymbol{\mu}^q,\boldsymbol{\sigma}^q)$ where $\boldsymbol{\mu}=(\mu_1,\mu_2,...,\mu_n)^T$ and $\boldsymbol{\sigma}=(\sigma_1,\sigma_2,...,\sigma_n)^T$,
\footnotesize
\begin{equation}
    D_{\rm KL}[q(\boldsymbol{z})\Vert p(\boldsymbol{z})]=\sum_{i=1}^n\left\{\ln\left(\frac{\sigma^p_i}{\sigma^q_i}\right)+\frac{(\mu^p_i-\mu^q_i)^2+(\sigma^q_i)^2}{2(\sigma^p_i)^2}-\frac{1}{2}\right\}
\end{equation}
\normalsize
The parameters of the model, including an adaptive variable $\boldsymbol{a}$ introduced in the following section, are optimized using BPTT. To perform error-regression explained in section \ref{error-regression}, an encoder was also trained separately.

\medskip
\subsection{Error-regression with shifting window} \label{error-regression}
\begin{figure}[ht]
    \centering
    \includegraphics[width=1.0\textwidth]{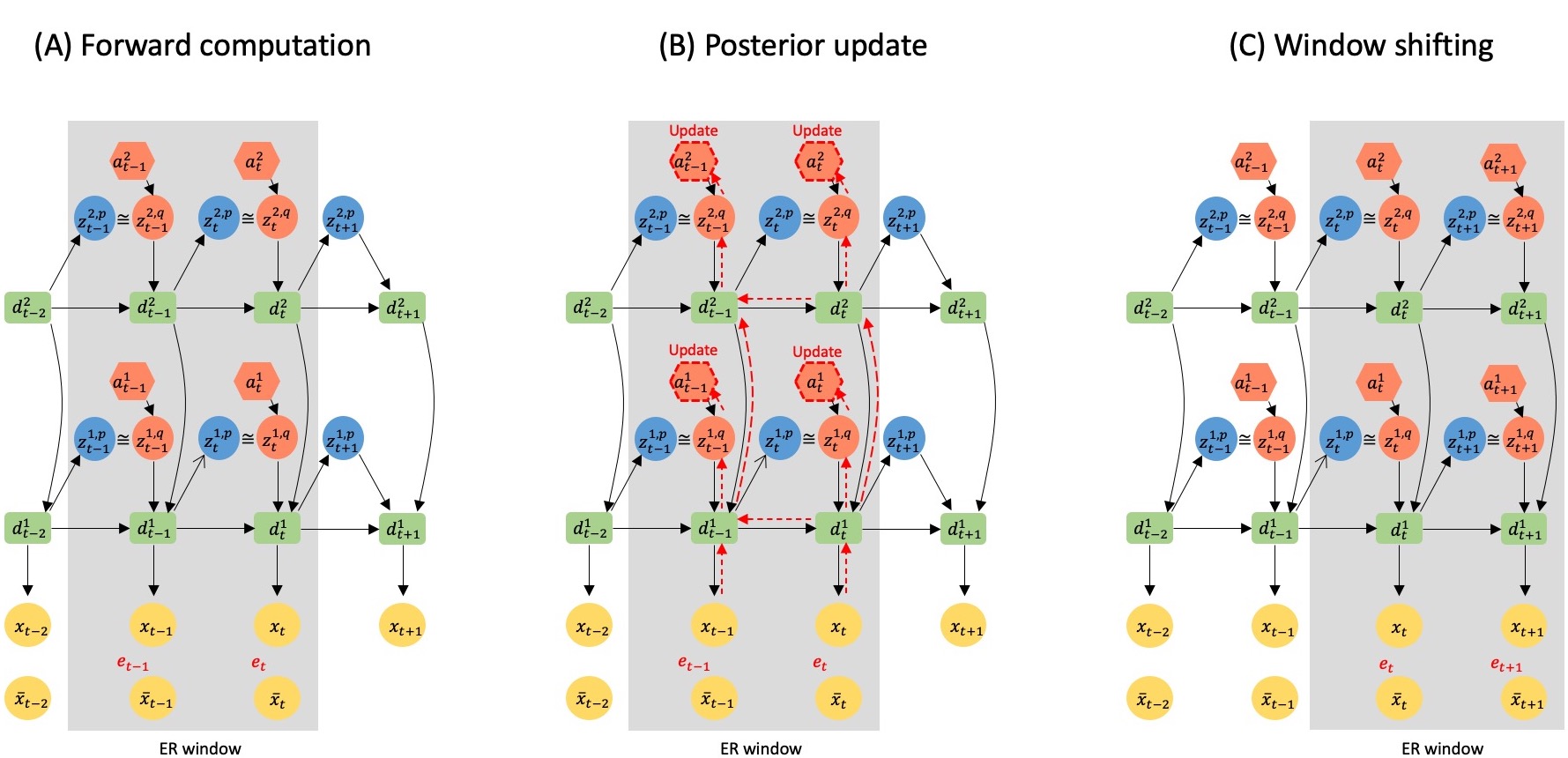}
    \caption{A graphical representation of error-regression with a shifting window. The gray area represents the ER window. Black arrows indicate forward computations. Red arrows indicate how reconstruction errors are propagated to $\boldsymbol{a}$ inside the ER window by BPTT. (A) illustrates the information flow of forward computation at time step $t$. (B) shows the update of the posterior inside the ER window at time step $t$. (C) shows the window shifting to time step $t+1$.}
    \label{er}
\end{figure}

\cite{ahmadi2019novel} proposed a scheme, the error-regression (ER) with shifting window to test the trained model in a way that is consistent with concepts of predictive coding and active inference. In this scheme, the trained network attempts to predict sensory inputs in the next time step while inferring the posterior in the immediate past window of a fixed length, using the reconstruction error in the window. The window is referred to as the ER window in the following. It is essential to note that ER for maximizing the evidence lower bound is conducted by iterating two processes of forward computation (Figure \ref{er} (A)) and posterior update (Figure \ref{er} (B)) for specific times at each sensory sampling time step. 

PV-RNN has unique variables $\boldsymbol{a}$ that facilitate updating the posterior. $\boldsymbol{a}$ is time step-specific and has the same dimension as $\boldsymbol{z}$ in each PV-RNN layer. In other words, when a PV-RNN layer with $\boldsymbol{z}$ with its dimensionality $n$ tries to infer the posterior for the last $T$ time steps inside the ER window, the PV-RNN layer has $n\times T\ \boldsymbol{a}$ valuables and updates them to modify the representation of the posterior. A detailed computation scheme of the posterior using the adaptive variable $\boldsymbol{a}$ is found in section \ref{model_implementation}. Importantly, in ER, weights and biases of the network are fixed, and only the adaptive variables $\boldsymbol{a}$ are updated. 

Let us consider an example of error-regression in which the length of the ER window is two time steps, and the network has two layers, as shown in Figure \ref{er}. Figure \ref{er} (A) illustrates the forward computation at time step $t$ to infer the posterior. In the forward computation, the network computes the conditional prior, $p(\boldsymbol{z_{t-1}}|\boldsymbol{d}_{t-2})$ and $p(\boldsymbol{z}_t|\boldsymbol{d}_{t-1})$, and the posterior, $q(\boldsymbol{z}_{t-1}|\boldsymbol{d}_{t-2}, e_{t-1:t})$ and $q(\boldsymbol{z}_{t}|\boldsymbol{d}_{t-1}, e_{t})$ in each layer, and generates the prediction with sampling from the posterior distribution inside the window. Then, the reconstruction error $e_{t-1}$ and $e_{t}$, and the KL divergence between respective pairs of the conditional prior and posterior are computed. 

Based on the reconstruction error and KL divergence, the inferred posterior is updated to maximize the evidence lower bound. Figure \ref{er} (B) illustrates how the reconstruction error is back-propagated through variables and layers to $\boldsymbol{a}$, which is responsible for updating the posterior. Using the updated posterior, the network again performs the forward computation. It should be noted that since the $q(\boldsymbol{z}_{t-1}|\boldsymbol{d}_{t-2}, e_{t-1:t})$ has been updated, $\boldsymbol{d}_{t-1}$ is different from the one before the update; thus, $p(\boldsymbol{z}|\boldsymbol{d}_{t-1})$ is also changed through the posterior update. The reconstruction error and the KL divergence are further computed, and the posterior is updated. This iterative process of forward computation and posterior update is repeated a fixed number of times to optimize the approximate posterior for maximizing the evidence lower bound computed with given meta-prior values.

After finishing all iterations, the network generates a new sensory prediction $\boldsymbol{x}_{t+1}$ with a conditional prior using the inferred posterior inside the ER window. Then the ER window shifts one time step and the next target sensation $\boldsymbol{x}_{t+1}$ is sampled, and the forward computation and the posterior update are reiterated at time step $t+1$. In the proposed model, the ER is performed for both visual sensation and proprioception simultaneously, and this scheme of using ER with a shifting window was used to test an imitative interaction after training the entire network, as will be described later.

\newpage
\subsection{Model implementation}\label{model_implementation}
The proposed model for the imitative interaction via visuo-proprioceptive sensation consists of three modules: an associative module, a proprioception module, and a vision module. This subsection describes a detailed computation scheme in each module.

\medskip
\subsubsection{The associative module}
The associative module is comprised of a PV-RNN. Since we adopted an MTRNN computation scheme in PV-RNN, its computations are as follows.
\postdisplaypenalty=0
\footnotesize
\begin{numcases}{\boldsymbol{u}_t^{a,l}=}
    \boldsymbol{W}_{dd}^{a,ll}\boldsymbol{d}_{t-1}^{a, l}+\boldsymbol{W}_{dz}^{a,ll}\boldsymbol{z}^{a,l}_t+\boldsymbol{b}^{a,l} & {\rm if\ top\ layer}\\
    \boldsymbol{W}_{dd}^{a,ll}\boldsymbol{d}_{t-1}^{a, l}+\boldsymbol{W}_{dd}^{a,ll+1}\boldsymbol{d}_t^{a,l+1}+\boldsymbol{W}_{dz}^{a,l}\boldsymbol{z}^{a,l}_t+\boldsymbol{b}^{a,l} & {\rm otherwise}
\end{numcases}
\begin{align}
    \boldsymbol{h}_t^{a,l}&=\left(1-\frac{1}{\tau^{a,l}}\right)\boldsymbol{h}_{t-1}^{a,l}+\frac{1}{\tau^{a,l}}\boldsymbol{u}_{t^l}\\
    \boldsymbol{d}_t^{a,l}&=\tanh\left(\boldsymbol{h}_t^{a,l}\right)
\end{align}
\normalsize
where $\boldsymbol{u}_t^{a,l}$ is the sum of inputs to $l$th layer of the associative module. $\boldsymbol{W}_{dd}^{a,ll}$, $\boldsymbol{W}_{dz}^{a,ll}$, and $\boldsymbol{W}_{dd}^{a,ll+1}$ are weight matrices for recurrent connections, the stochastic variable $\boldsymbol{z}$, and the input from the higher layer, respectively. $\boldsymbol{b}^{a,l}$ is the bias in the $l$th layer in the associative module, and $\tau^{a,l}$ is the time constant for MTRNN computation in the $l$th layer of the associative module. $\tanh$ is the activation function. The stochastic variable $\boldsymbol{z}$ is assumed to follow a multivariate Gaussian distribution with a diagonal covariant matrix, and the deterministic variable $\boldsymbol{d}$ predicts mean $\boldsymbol{\mu}$ and variance $\boldsymbol{\sigma}$ of the distribution. That is, for computation of the prior,
\postdisplaypenalty=0
\footnotesize
\begin{numcases}{p(\boldsymbol{z}_t^{p,a,l})=}
    \mathcal{N}(\boldsymbol{z}^u;\boldsymbol{0},\boldsymbol{I}) & {\rm if\ } t=1\\
    p(\boldsymbol{z}_t^{p,a,l}|\boldsymbol{d}_{t-1}^{a,l})=\mathcal{N}(\boldsymbol{z}_t^{p,a,l};\boldsymbol{\mu}_t^{p,a,l},\boldsymbol{\sigma}_t^{p,a,l}) & {\rm otherwise} 
\end{numcases}
\begin{align}
    &\boldsymbol{\mu}_t^{p,a,l}=\tanh(\boldsymbol{W}_{\mu d}^{a,l}\boldsymbol{d}_{t-1}^{a,l} + \boldsymbol{b}_\mu^{a,l})\\
    &\boldsymbol{\sigma}_t^{p,a,l}=\exp(\boldsymbol{W}_{\sigma d}^{a,l}\boldsymbol{d}_{t-1}^{a,l}+\boldsymbol{b}_\sigma^{a,l})\\
    &\boldsymbol{z}_t^{p,a,l}=\boldsymbol{\mu}_t^{p,a,l}+\boldsymbol{\sigma}_t^{p,a,l}*\boldsymbol{\epsilon}
\end{align}
\normalsize
where $\boldsymbol{\mu}_t^{p,a,l}$ and $\boldsymbol{\sigma}_t^{p,a,l}$ are the mean and variance for the prior distribution of $\boldsymbol{z}_t^{p,a,l}$ at time step $t$ in $l$th layer in the associative module. $\boldsymbol{W}_{\mu d}^{a,l}$ and $\boldsymbol{W}_{\sigma d}^{a,l}$ are the weight matrices for $\boldsymbol{d}^{a,l}_{t-1}$. $\boldsymbol{b}_{\mu}^{a,l}$ and $\boldsymbol{b}_{\sigma}^{a,l}$ are the biases for each computation. $\tanh$ in computation of mean is used for stability of optimization, and $\exp$ in $\boldsymbol{\sigma}$ is for variance to be positive. $\boldsymbol{\epsilon}$ is sampled from $\mathcal{N}(\boldsymbol{0},\boldsymbol{I})$. To approximate the posterior, PV-RNN has adaptive variables $\boldsymbol{a}$ that are specific to time step and sequence. $\boldsymbol{a}$ is optimized during learning with the prediction error via BPTT. By considering $\boldsymbol{a}$, the computations of posterior are
\postdisplaypenalty=0
\footnotesize
\begin{align}
    &q(\boldsymbol{z}_t^{q,a,l}|\boldsymbol{d}_{t-1}^{a,l}, e_{t:T}^p, e_{t:T}^v)=\mathcal{N}(\boldsymbol{z}_t^{q,a,l};\boldsymbol{\mu}_t^{q,a,l},\boldsymbol{\sigma}_t^{q,a,l})\\
    &\boldsymbol{\mu}_t^{q,a,l}=\tanh(\boldsymbol{W}_{\mu d}^{a,l}\boldsymbol{d}_{t-1}^{a,l} + \boldsymbol{a}_{\mu,t}^{a,l} + \boldsymbol{b}_\mu^{a,l})\\
    &\boldsymbol{\sigma}_t^{q,a,l}=\exp(\boldsymbol{W}_{\sigma d}^{a,l}\boldsymbol{d}_{t-1}^{a,l}+\boldsymbol{a}_{\sigma,t}^{a,l}+\boldsymbol{b}_\sigma^{a,l})\\
    &\boldsymbol{z}_t^{q,a,l}=\boldsymbol{\mu}_t^{q,a,l}+\boldsymbol{\sigma}_t^{q,a,l}*\boldsymbol{\epsilon}
\end{align}
\normalsize
where $\boldsymbol{\mu}_t^{q,a,l}$ and $\boldsymbol{\sigma}_t^{q,a,l}$ are mean and variance for the posterior distribution of $\boldsymbol{z}_t^{q,a,l}$ at time step $t$ in $l$th layer in the associative module. Note that the weight matrices for $\boldsymbol{d}$ are different from those used to compute the prior. In addition, unlike the peripheral sensory modules of proprioception and vision, the associative module does not predict the sensory output directly, but rather predicts the latent representation of visuo-proprioceptive sequences. Therefore, the weights and biases, as well as the adaptive variable $\boldsymbol{a}$ of the associative module are not optimized instantly from the reconstruction error of the sensory outcomes, but from the error signals mediated through each sensory module. 

\medskip
\subsubsection{The proprioception module}\label{prop}
Proprioceptive patterns are directly generated from the PV-RNN. The highest layer in the proprioception module receives the input from the lowest layer in the associative layer, and its computations are
\footnotesize
\postdisplaypenalty=0
\begin{numcases}{\boldsymbol{u}_t^{p,l}=}
    \boldsymbol{W}_{dd}^{pa}\boldsymbol{d}_t^{a,1}+\boldsymbol{W}_{dd}^{p,ll}\boldsymbol{d}_{t-1}^{p,l}+\boldsymbol{W}_{dz}^{p,l}\boldsymbol{z}_t^{p,l}+\boldsymbol{b}^{p,l} & {\rm if\ top\ layer}\\
    \boldsymbol{W}_{dd}^{p,ll}\boldsymbol{d}_{t-1}^{p,l}+\boldsymbol{W}_{dz}^{p,l}\boldsymbol{z}_t^{p,l}+\boldsymbol{b}^{p,l} & {\rm otherwise}
\end{numcases}

\begin{align}
    \boldsymbol{h}_t^{p,l}&=\left(1-\frac{1}{\tau^{p,l}}\right)\boldsymbol{h}_{t-1}^{p,l}+\frac{1}{\tau^{p,l}}\boldsymbol{u}_t^{p,l}\\
    \boldsymbol{d}_t^{p,l}&=\tanh\left(\boldsymbol{h}_t^{p,l}\right)
\end{align}
\normalsize
A proprioceptive pattern at time step $t$, $\boldsymbol{p}_t$, is generated from the lowest layer of the proprioception module.
\footnotesize
\begin{equation}
    \boldsymbol{p}_t=\tanh\left(\boldsymbol{W}^p\boldsymbol{d}_t^{p,1}+\boldsymbol{b}^p\right)
\end{equation}
\normalsize

\medskip
\subsubsection{The vision module}\label{vision_module}
For the vision module, a scheme to reduce the computation time is introduced. As described in section \ref{error-regression} above, in the proposed imitative interaction scheme, the network is required to infer the posterior for the immediate past at every sensory sampling time step by repeating forward computation and BPTT, which demands intensive computation. Nevertheless, our model is expected to work in actual robots in real-time in the future, which necessitates reducing the model's computational complexity. To reduce the computational demand in the posterior inference in visual perception, we consider a composite network combining a dynamic PV-RNN and static CNNs for decoding and encoding pixel patterns, instead of introducing full recurrent connections in this module. In this composite network, when generating predictive output for the visual input, the PV-RNN part predicts the latent state representation with a relatively low dimension, which is fed to a CNN decoder to generate the corresponding visual pixel image. On the other hand, when receiving the visual input, it is transformed to the latent state representation by a CNN encoder. Then, the prediction error can be computed as the discrepancy in the latent state with a low dimension rather than at the pixel level with high dimension. This reduces the computational burden significantly for conducting the BPTT to infer the posterior during imitative interaction. As in the proprioception module, the highest layer of the vision module receives input from the lowest layer of the associative layer, and its computations are
\postdisplaypenalty=0
\footnotesize
\begin{numcases}{\boldsymbol{u}_t^{v,l}=}
    \boldsymbol{W}_{dd}^{va}\boldsymbol{d}_t^{a,1}+\boldsymbol{W}_{dd}^{v,ll}\boldsymbol{d}_{t-1}^{p,l}+\boldsymbol{W}_{dz}^{v,l}\boldsymbol{z}_t^{v,l}+\boldsymbol{b}^{v,l} & {\rm if\ top\ layer}\\
    \boldsymbol{W}_{dd}^{v,ll}\boldsymbol{d}_{t-1}^{v,l}+\boldsymbol{W}_{zd}^{v,l}\boldsymbol{z}_t^{v,l}+\boldsymbol{b}^{v,l} & {\rm otherwise}
\end{numcases}
\begin{align}
    \boldsymbol{h}_t^{v,l}&=\left(1-\frac{1}{\tau^{v,l}}\right)\boldsymbol{h}_{t-1}^{v,l}+\frac{1}{\tau^{v,l}}\boldsymbol{u}_t^{v,l}\\
    \boldsymbol{d}_t^{v,l}&=\tanh\left(\boldsymbol{h}_t^{v,l}\right)
\end{align}
\normalsize
Then the lowest layer of the PV-RNN predicts the latent state $\boldsymbol{l}_t$ at time step $t$, and the visual pattern $\boldsymbol{v}_t$ is generated by the decoder.
\footnotesize
\postdisplaypenalty=0
\begin{align}
    &\boldsymbol{l}_t=\tanh\left(\boldsymbol{W}^l\boldsymbol{d}_t^{v,1}+\boldsymbol{b}^l\right)\\
    &\boldsymbol{v}_t={\rm decoder}(\boldsymbol{l}_t)
\end{align}
\normalsize
In the imitative interaction, the target of latent dynamics $\bar{\boldsymbol{l}_t}$ of visual patterns $\bar{\boldsymbol{v}_t}$ at time step $t$ is computed by the encoder.
\footnotesize
\begin{equation}
    \bar{\boldsymbol{l}}_t={\rm encoder}(\bar{\boldsymbol{v}}_t)
\end{equation}
\normalsize
To improve the generalization capability of the encoder and decoder, {\it CoordConv} architecture\citep{liu2018intriguing} was introduced.

\section{Experiments}
\subsection{Experimental design}   
Using the proposed model, imitative interaction experiments considering human-robot interactions were conducted. Although human-robot interactions ought to be studied in an online fashion to reflect human behavior in response to robot actions, because of the intensive computation required in the error regression scheme, we could not conduct such experiments online. Therefore, the current study examined only the dynamic response of the model network using recorded sequences of visuo-proprioceptive patterns. Therefore, data containing human-demonstrated movement patterns in terms of visuo-proprioceptive sequences were collected both for training the network and for later testing of pseudo-synchronized imitative interaction. After training, the model was tested for pseudo-imitative interaction using novel visuo-proprioceptive patterns with two different scenarios (Experiment 1 and Experiment 2). 

Experiment 1 investigated the issue of coordination and integration of different modalities of sensation by changing the tightness used to regulate the complexity term for each sensory module. For this purpose, the network was trained by assigning different values of the meta-prior to the proprioception and vision modules. We examined the different effects of regulating complexity in the two modules on coordination of different modalities by analyzing them in both the learning process and in the pseudo-imitative interaction tested after learning. 

Experiment 2 investigated the issue of strength of agency as the main motivation of the current study by changing the tightness used to regulate the complexity term for the entire network from that introduced in the training phase. Accordingly, we selected a network trained and evaluated as successful in Experiment 1 and then the characteristics of the pseudo-imitative interaction were examined by equally adjusting the meta-priors of each module of this trained network to larger or smaller values.

In subsequent experiments, some parameters that determine network structure were set as follows. The associative module consisted of a one-layer PV-RNN, and the proprioception module and the vision module consisted of two layers. These PV-RNN layers were characterized by a time scale imposed on MTRNN computation. That is, the higher layer had a larger time constant, producing slow time-scale dynamics, and the lower layer had smaller time constants, generating fast time-scale dynamics. Therefore, in this study, the higher layer of the proprioception module and the vision module, which receive input from the associative module, are referred to as the {\it slow layer}, and the lower layer is referred to as the {\it fast layer}. As described in section \ref{vision_module}, the visual perception of the model involves an encoder and a decoder. Their architectures are summarized in Table \ref{enc_dec}.

\begin{table}[ht]
    \centering
    \begin{tabular}{lcccc}\hlineB{1.5}
    Layer & Kernel size & Stride & Filter & Activation\\
    \hline\\
    \multicolumn{5}{c}{Encoder}\\
    \hline\hline\\
    Conv & 33$\times$33 & 1 & 5 & ReLu \\
    Conv & 17$\times$17 & 1 & 15 & ReLu \\
    Conv & 16$\times$16 & 1 & 30 & $\tanh$ \\
    \hline\\
    \multicolumn{5}{c}{Decoder}\\
    \hline\hline\\
    Conv transpose & 16$\times$16 & 1 & 15 & ReLu\\
    Conv transpose & 17$\times$17 & 1 & 5 & ReLu\\
    Conv transpose & 33$\times$33 & 1 & 1 & $\tanh$\\\hlineB{1.5}
    \end{tabular}
    \caption{The architecture of the encoder and the decoder.}
    \label{enc_dec}
\end{table}

\medskip
\subsection{Data preparation}\label{data preparation}
\begin{figure}[ht]
    \centering
    \includegraphics[width=1.0\textwidth]{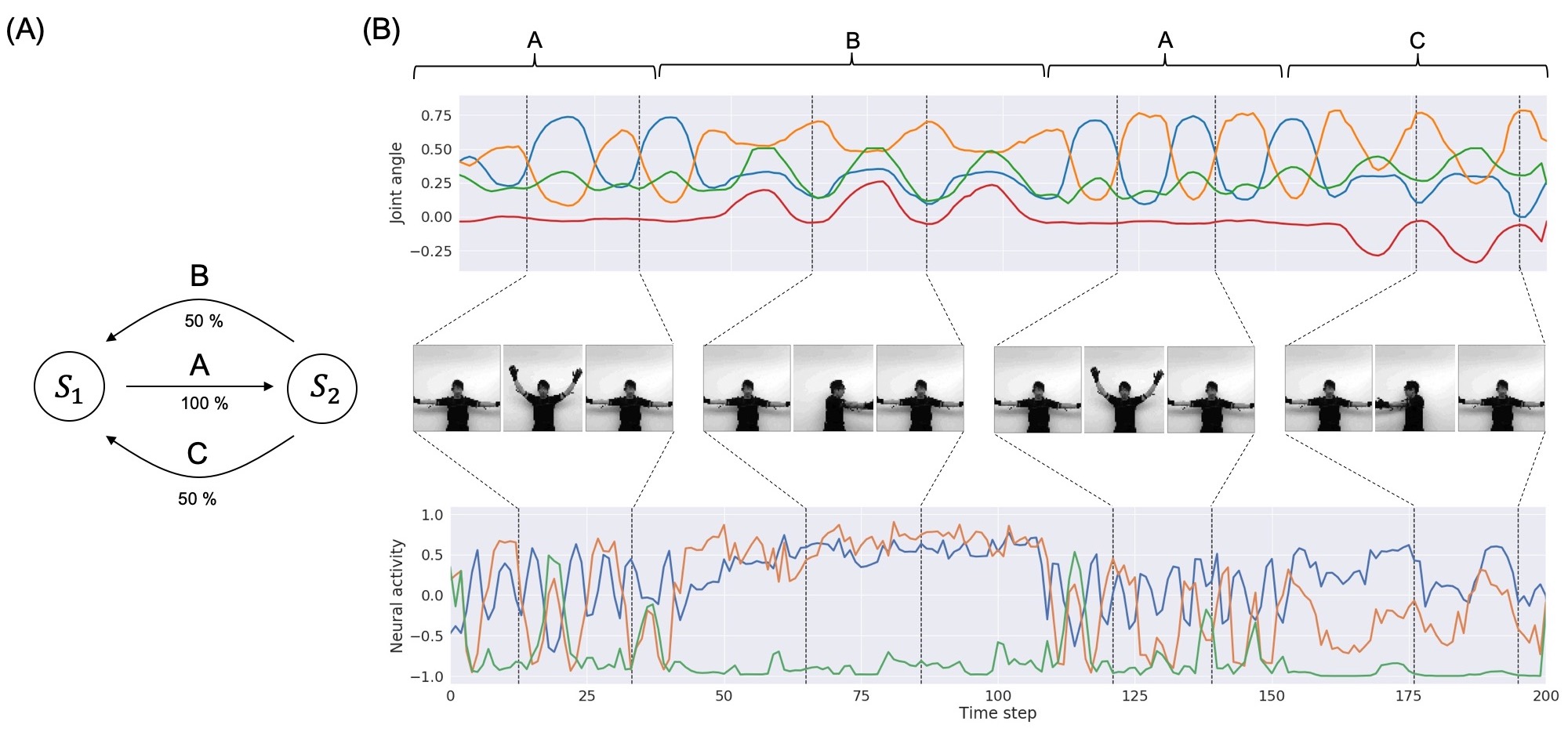}
    \caption{Training data. (A) A diagram of the probabilistic finite state machine. (B) An example of the training dataset. The top row is part of a joint-angle trajectory. The corresponding labels of primitive patterns (A, B, and C) are indicated above the plots. For simplicity, only 4 joint-angles out of 16 representing the movements are shown. The middle row shows corresponding visual pixel images in each period. The bottom row shows visual trajectories in the latent space embedded by the encoder. For simplicity, only three variables out of 20 are shown.}
    \label{training_example}
\end{figure}

To obtain a dataset of synchronized visuo-proprioceptive sequences, we used a humanoid robot, Torobo (Tokyo Robotics Inc.) and a motion capture suit (Perception Neuron, Noitom Ltd.). Torobo is a human-sized, torso-type humanoid robot with 16 joint-angles, of which 6 are for each arm and 4 are for the torso and head positions. Human body movements can be mapped to joint-angle trajectories of the robot using the motion capture suit. A human experimenter wearing the suit demonstrated a set of body movements, which were mapped as joint-angle trajectories. This demonstration was also recorded with a camera to obtain corresponding visual patterns. The target sequential movement pattern to be learned by the robot was designed by considering a probabilistic finite state machine that can generate probabilistic sequences of three different primitive movement patterns. Those were (A) waving with both arms three times, (B) rotating the torso to the left with the arms three times, and (C) rotating the torso to the right with the arms three times. Primitive pattern A is followed either by primitive pattern B or primitive pattern C with a 50\% chance, and primitive patterns B and C are followed by pattern A with a 100\% chance (Figure \ref{training_example} (A)). One sequence consists of 8 probabilistic transitions of primitive movements. Three human participants demonstrated and recorded 10 movement sequences each. In other words, the dataset comprised 30 sequences of visuo-proprioceptive temporal patterns. Recorded visuo-proprioceptive patterns were down-sampled to 3.75 Hz so that one sequence became 400 time steps. Joint-angle trajectories were normalized to a range between $-1$ and $1$. Vision patterns were further converted into gray scale images and down-sized to $64\times64$ pixels (Figure \ref{training_example} (B)). A summary of the training data is shown in Table \ref{training_data}. Visual trajectories fluctuated far more than proprioceptive trajectories due to various optical conditions, such as illumination and surface reflectiveness.

\begin{table}[ht]
    \centering
    \begin{tabular}{ccccc}\hlineB{1.5}
         & Dimension & time step & Participants & Total sequences \\ \hline\\
         Proprioception & 16 & \multirow{2}{*}{400} & \multirow{2}{*}{3} & \multirow{2}{*}{30}\\
         Vision & 64$\times$64\\\hlineB{1.5}
    \end{tabular}
    \caption{A summary of the training data.}
    \label{training_data}
\end{table}

Using the training example, the model is required to extract a probabilistic structure such that the primitive pattern of B or C appears with only a 50\% chance after every appearance of the primitive A, by estimating precision in transitions of primitives with learning.
Such learning should be achieved without providing explicit labels for those primitives, by extracting the underlying chunking and segmentation structure from continuous sensory signals prepared in the dataset. The PV-RNN can achieve such tasks using a multiple timescale RNN scheme combined with a Bayesian inference approach \citep{ahmadi2019novel}.

\medskip
\subsection{Experiment 1: Training with different meta-priors in different modalities}
This experiment investigates effects of changing the tightness used to regulate the complexity term for each sensory module with regard to coordination and integration of different modalities of sensation. 
In addition, this experiment provides successfully trained networks with well-balanced complexity between the vision and proprioceptive modules for possible use in Experiment 2.
To accomplish this, we examined how assigning different values of the meta-prior to the proprioception and vision modules affects the learning process and performance in the pseudo-imitative interaction. Two sets of meta-priors $w_1$ and $w_2$ were assigned to the model (Table \ref{network configuration}). $w_1$ has larger values of the meta-prior in the proprioception module than in the vision module, and they were exchanged in $w_2$. Both $w_1$ and $w_2$ have the same value for the meta-prior in the associative module. First, the model was trained with the $w_1$ and $w_2$ settings, and the learning process was examined, with special attention to each component of the lower bound. To facilitate training, the Adam optimizer \citep{kingma2014adam} was utilized with the parameter settings $\alpha=0.001$, $\beta_1=0.9$, and $\beta_2=0.999$. The model was trained 10 times with different random initializations of model parameters for 10,000 epochs, and the mean and standard deviation of the prediction errors of proprioception and vision, and the KL divergence of each layer of the model at each epoch were computed.

\begin{figure}[ht]
    \centering
    \includegraphics[width=1.0\textwidth]{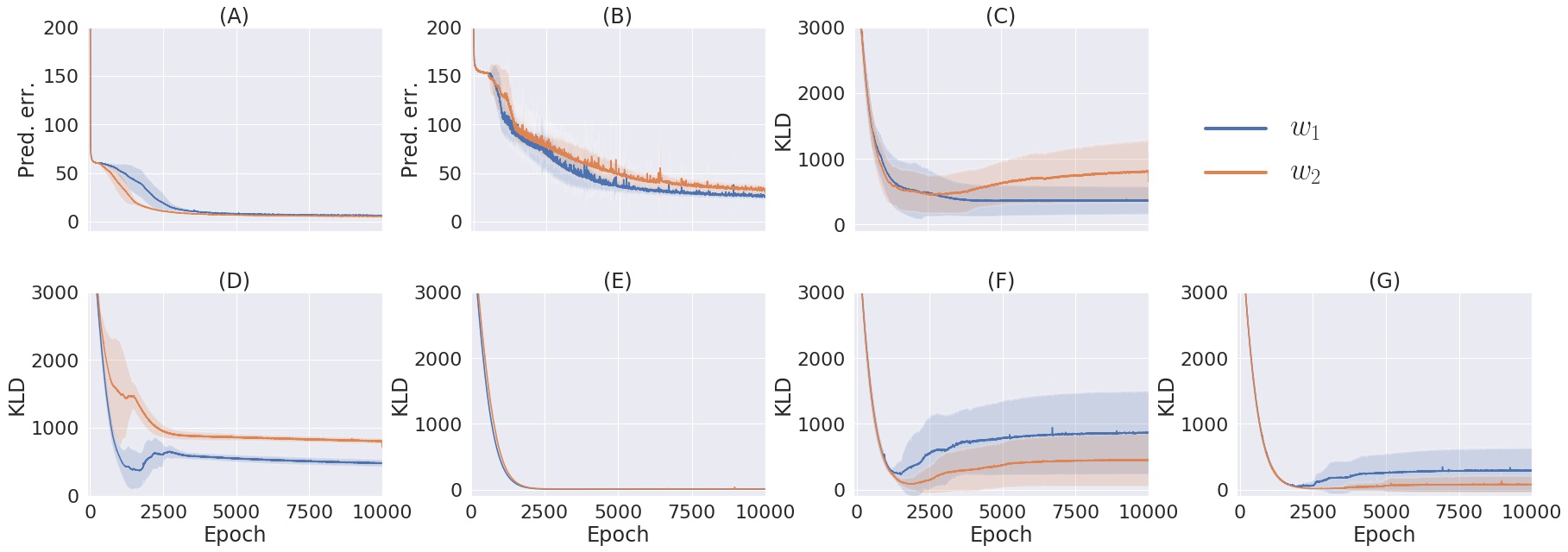}
    \caption{The learning process of the model with two different meta-prior settings. (A) The prediction error in proprioception. (B) The prediction error in vision. (C) The KL divergence in the associative module. (D) The KL divergence in the slow layer of the proprioception module. (E) The KL divergence in the fast layer of the proprioception module. (F) The KL divergence in the slow layer of the vision module. (G) The KL divergence in the fast layer of the vision module. The shadows are the standard deviation of 10 trials with different parameter initializations. Note that values of prediction errors are the sum of the prediction errors at all time steps and sequences normalized by the data dimension.}
    \label{ex1_1}
\end{figure}

Results are summarized in Figure \ref{ex1_1}. In comparing $w_1$ and $w_2$ conditions, even though the prediction errors in the proprioception and vision modules showed similar behavior (Figure \ref{ex1_1} (A), (B)), the KL divergence in each module was optimized differently. Despite different values of the meta-prior assigned to the fast layer of the proprioception module, its KL divergences in $w_1$ and $w_2$ conditions were reduced in exactly the same way (Figure \ref{ex1_1} (E)). This is not the case in the fast layer of the vision module (Figure \ref{ex1_1} (G)). The KL divergence in the slow layer of the proproception module and the slow layer of the vision module showed different values in $w_1$ and $w_2$ settings (Figure \ref{ex1_1} (D),(F)). Interestingly, although the associative module was set to the same value of meta-prior in $w_1$ and $w_2$ conditions, the KL divergence in the $w_2$ setting reached a larger value than in the $w_1$ setting. This is because the larger value of the meta-prior assigned to the fast layer of the vision module in the $w_2$ condition prevented the vision module from absorbing the fluctuation in observed visual patterns, which resulted in bottom-up fluctuation from the vision module to the associative module, appearing as a discrepancy between the prior and the posterior in this module. Because visual sensation contains more inherent randomness than proprioceptive sensation, as mentioned previously, complexity in this modality should be adequately regulated by setting a smaller meta-prior value. Otherwise, the discrepancy that appears in the visual module tends to leap to the higher associative module without being well resolved before.

\begin{table}[h]
\centering
\begin{tabular}{lccccccc}\hlineB{1.5}
            & \multirow{2}{*}{$\mathbb{R}^d$} & \multirow{2}{*}{$\mathbb{R}^z$} & \multirow{2}{*}{$\tau$} & \multicolumn{2}{c}{$w_1$ setting} & \multicolumn{2}{c}{$w_2$ setting}  \\ \cline{5-6} \cline{7-8}
            &   &   &   &  $w^l$  &  $w_1^l$  &  $w^l$  &  $w_1^l$ \\ \hline\hline\\
Assoc. module     &  10  &  1  &  15  &  0.0025  &  0.01  &  0.0025  &  0.01  \\
Prop. slow layer  &  20  &  2  &  8   &  0.005   &  0.01  &  0.0025  &  0.05  \\
Prop. fast layer  &  30  &  3  &  2   &  0.01    &  0.01  &  0.005   &  0.05  \\
Vision slow layer &  20  &  2  &  8   &  0.0025  &  0.05  &  0.005   &  0.01  \\
Vision fast layer &  30  &  3  &  2   &  0.005   &  0.05  &  0.01    &  0.01  \\\hlineB{1.5}             
\end{tabular}
\caption{The model configuration in Experiment 1. $\mathbb{R}^d$ and $\mathbb{R}^z$ are the dimensions of $\boldsymbol{d}$ and $\boldsymbol{z}$, respectively. $\tau$ is the time constant of the MTRNN computation in each layer.}
\label{network configuration}
\end{table}

\medskip
We further tested the trained models in the pseudo-imitative interaction. Training of the models stopped after 4,000 epochs. Three novel visuo-proprioceptive sequences recorded from three human participants were prepared for the pseudo-imitative interaction, which also comprised the previous primitive body movements A, B, and C, the lengths of which were 400 time steps. The length of the ER window was set to 30 time steps, and the number of optimization iterations for posterior inference by BPTT at each time step was 30. Namely, at each sensory sampling time step, the network infers the posterior distribution of $\boldsymbol{z}$ responsible for reconstructing the observation inside the ER window, in which the cycle of the forward computation and the posterior update described in section \ref{error-regression} repeats 30 times. As in learning, Adam was used to improve optimization with parameter settings $\alpha=0.2$, $\beta_1=0.9$, and $\beta_2=0.999$. Evaluation of the error-regression examined how much the reconstruction error in each modality and the KL divergence at each sub-network in the PV-RNN were minimized. That is, at the point when $T'$ time step window for the immediate past shifts $t$ time steps, i.e., the current time step is $t$, the adaptive variable $\boldsymbol{a}$ assigned within the window is optimized with the iterative process, and at the last iteration, the reconstruction error and the KL divergence are computed inside the window. Therefore, they are defined as
\footnotesize
\postdisplaypenalty=0
\begin{align}
    &{\rm Proprioception\ error}:=\frac{1}{T}\sum_{t=1}^{T}\frac{1}{T'}\sum_{t'=1}^{T'}\frac{1}{R^p}\Vert \boldsymbol{p}_{t'}-\bar{\boldsymbol{p}}_{t'}\Vert^2\\
    &{\rm Vision\ error} := \frac{1}{T}\sum_{t=1}^T\frac{1}{T'}\sum_{t'=1}^{T'}\frac{1}{R^l}\Vert \boldsymbol{l}_{t'}-\bar{\boldsymbol{l}}_{t'}\Vert^2\\
    &{\rm KLD} := \frac{1}{T}\sum_{t=1}^{T}\frac{1}{T'}\sum_{t'=1}^{T'}\frac{1}{R^z}D_{\rm KL}[q(\boldsymbol{z}_{t'}|\boldsymbol{d}_{t'-1},e_{t':T})\Vert p(\boldsymbol{z}_{t'}|\boldsymbol{d}_{t'-1})]
\end{align}
\normalsize
where  $t'$ is the time step inside the window. $R^p$ and $R^l$ are the dimension of proprioception and the latent space of vision, respectively. $R^z$ is the dimension of $\boldsymbol{z}$, and the KL divergence is computed for every PV-RNN submodule. Models trained in previous experiments were used. The pseudo-imitative interaction experiment was run 10 times with different random number seeds, and the mean and standard deviation of each quantity were computed. In addition, one-step, look-ahead prediction error, the discrepancy between the prediction in the next time step of the current window and the observation, was computed in the vision module to evaluate prediction accuracy.

\begin{figure}[ht]
    \centering
    \includegraphics[width=1.0\textwidth]{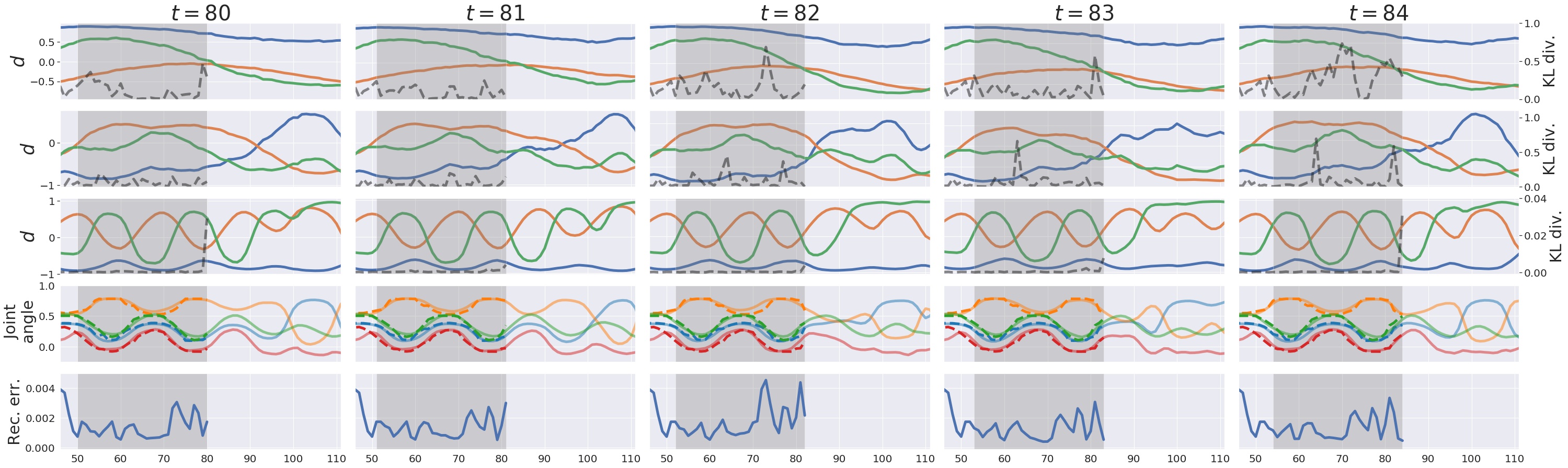}
    \caption{An example of the network representation during testing in $w_1$ setting. Gray areas indicate the ER window. The first, second, and third rows show representations in the associative module, the slow layer in the proprioception module, and the fast layer in the proprioception module, respectively. Solid lines represent activities of three randomly chosen $\boldsymbol{d}$ neurons, and dashed lines represent the value of the KL divergence at each time step. The 4th row shows predictions (solid lines) and sensations (dashed lines) of joint-angles. For clarity, only four joint-angles of 16 are shown. The bottom row shows the reconstruction error in proprioception.}
    \label{exp1_2_1}
\end{figure}

Figure \ref{exp1_2_1} exemplifies how the pseudo-imitative interaction developed in the $w1$ setting in time-lapse. For clarity, only parts involving the proprioceptive interaction are shown. Each column shows the representation of the network when the network finished a posterior inference and made a new prediction at each time step. The first, second, and third row show representations in the associative module, the slow layer in the proprioception module and the fast layer in the proprioception module, respectively. Solid lines indicate the activity of three randomly chosen $\boldsymbol{d}$ neurons, and dashed lines indicate the KL divergence value at each time step in each layer. The fourth row shows joint-angle trajectories. Solid lines are predictions generated by the network, and dashed lines are joint-angle values demonstrated by the human counterpart in the recorded data. The bottom row shows the reconstruction error, inside the ER window, which was minimized by updating $\boldsymbol{a}$ via BPTT under regularization by the KL divergence between the inferred posterior and the conditional prior. In section \ref{error-regression}, describing the error-regression scheme, the network is illustrated in a way that it only makes the prediction at next time step during the interaction. In this experiment, however, the network was allowed to generate the prediction not only at next time step, but also at subsequent time steps with the conditional prior to visualize the network's long-term prediction. This is also the case in Figure \ref{exp2-2}.

At each time step, the network receives a new sensation, computes the reconstruction error and the KL divergence within the ER window, updates the $\boldsymbol{a}$ such that the lower bound inside the ER window is maximized, and modifies the prediction after the current time step with the conditional prior. In Figure \ref{exp1_2_1}, the network continually modified the future prediction as a result of the posterior inference. Since the lower bound summed over time steps inside the ER window is maximized, all $\boldsymbol{a}$s inside the ER window are updated so that the sum of the reconstruction error and the KL divergence weighted by the meta-prior inside the ER window are minimized. Therefore, it is often observed that the value of the reconstruction error or the KL divergence at a certain time step inside the ER window becomes larger at the next time step, which is considered a transient process in the optimization wherein the past is re-interpreted and re-represented in coping with a new entering sensation, in terms of post-diction \citep{shimojo2014postdiction}. In the $w_1$ setting, larger values of the meta-prior are assigned to lower layers of the network and smaller to higher. In other words, KL divergences in lower layers are weighted more in the lower bound, and while those in higher layers are weighted less. Therefore, KL divergences in lower layers were reduced more, and those in higher layers remained larger after iterative optimization. Owing to MTRNN characteristics of different time-scales among layers, higher layers showed slower dynamics and lower layers showed faster dynamics.
It is assumed that higher levels predict switching of primitives and lower levels predict sensory profile changes at each time step. Detailed analysis of this issue was not conducted in the current study since similar phenomena using MTRNN have been reported frequently (e.g. \cite{yamashita2008, hwang2020dealing}).

Experimental results are summarized in Table \ref{exp1_2}. The reconstruction error in proprioception was remarkably minimized compared to that in vision, in both conditions $w_1$ and $w_2$. This is because vision involves more noise than proprioception. The reconstruction error in vision was smaller for the $w_1$ condition than the $w_2$ condition. Furthermore, the KL divergence in the associative module was reduced more significantly in the $w_1$ condition than the $w_2$ condition. This occurred because the vision module generalized better with noisy visual patterns in the test of pseudo-imitative interaction in the $w_1$ case than in the $w_2$ case by minimizing the complexity term more. Because fluctuation or randomness in visual sensation was well resolved in the vision module in the $w_1$ case, the associative module became relatively free from such fluctuation, as evidenced by the smaller KL divergence observed in the associative module. As a result, the one-step, look-ahead prediction was also more accurate.

\begin{table}[h]
    \centering
    \begin{tabular}{lccccc}\hlineB{1.5}
         & Proprioception & Vision & Associative & Proprioception\\
         & reconstruction error & reconstruction error & KLD & slow KLD\\
         \hline\hline\\
        $w_1$ & $0.017\pm9.5\times10^{-4}$ & $0.12\pm6.9\times10^{-3}$ & $2.0\pm0.091$ & $1.6\pm0.12$\\
        $w_2$ & $0.011\pm2.9\times10^{-4}$ & $0.19\pm1.5\times10^{-2}$ & $3.2\pm0.15$ & $2.6\pm0.11$\\
        \hline\\
         & Proprioception & Vision & Vision & Vision one-step\\
         & fast KLD & slow KLD & fast KLD & prediction error\\
        \hline\hline\\
        $w_1$ & $0.57\pm0.040$ & $8.1\pm0.17$ & $0.50\pm0.048$ & $0.20\pm0.0097$\\
        $w_2$ & $0.57\pm0.041$ & $1.9\pm0.071$ & $0.54\pm0.048$ & $0.24\pm0.017$\\\hlineB{1.5}
    \end{tabular}
    \caption{The result of the pseudo-imitative interaction experiment. The errors are the standard deviation of 10 different trials with different random number seeds.}
    \label{exp1_2}
\end{table}

\medskip
\subsection{Experiment 2: Imitation with stronger and weaker agency}
This experiment was devised to reveal possible effects of changing the tightness used to regulate the complexity term for the entire network on the strength of agency exerted in imitative interaction. Accordingly, we investigated how changes of meta-prior values of the entire network from default values used in learning affect performance characteristics in the pseudo-imitative interaction. We used a network that was trained for 4,000 epochs in Experiment 1 with the $w_1$ setting as the default network. Five meta-prior settings were prepared for testing of imitative interaction: from smaller values of the meta-prior setting $W_1$ to the larger setting $W_5$ with a consistent ratio among all layers of all modules (Table \ref{w setting in exp2}). Imitative interaction with different meta-prior settings was performed with the novel visuo-proprioceptive patterns used in Experiment 1. Interactions were analyzed in terms of the quantities introduced in previous experiments. In addition, one-step look-ahead prediction error in proprioception was also measured. Each test with a different meta-prior setting was repeated with 10 network models trained with different initialization weights, but with the same parameters for the purpose of examining these quantities statistically.

\begin{table}[h]
    \centering
    \begin{tabular}{cccccc}\hlineB{1.5}
         & $W_1$ & $W_2$ & $W_3$ & $W_4$ & $W_5$\\
         \hline\hline\\
        Associative module & $2.5\times10^{-5}$ & $2.5\times10^{-4}$ & $2.5\times10^{-3}$ & $2.5\times10^{-2}$ & $2.5\times10^{-1}$\\
        Proprioception slow layer & $5.0\times10^{-5}$ & $5.0\times10^{-4}$ & $5.0\times10^{-3}$ & $5.0\times10^{-2}$ & $5.0\times10^{-1}$\\
        Proprioception fast layer & $1.0\times10^{-4}$ & $1.0\times10^{-3}$ & $1.0\times10^{-2}$ & $1.0\times10^{-1}$ & $1.0$\\
        Vision slow layer &$2.5\times10^{-5}$ & $2.5\times10^{-4}$ & $2.5\times10^{-3}$ & $2.5\times10^{-2}$ & $2.5\times10^{-1}$\\
        Vision fast layer & $5.0\times10^{-5}$ & $5.0\times10^{-4}$ & $5.0\times10^{-3}$ & $5.0\times10^{-2}$ & $5.0\times10^{-1}$\\\hlineB{1.5}
    \end{tabular}
    \caption{The values of meta-prior in Experiment 2.}
    \label{w setting in exp2}
\end{table}

\begin{figure}[ht]
    \centering
    \includegraphics[width=1.0\textwidth]{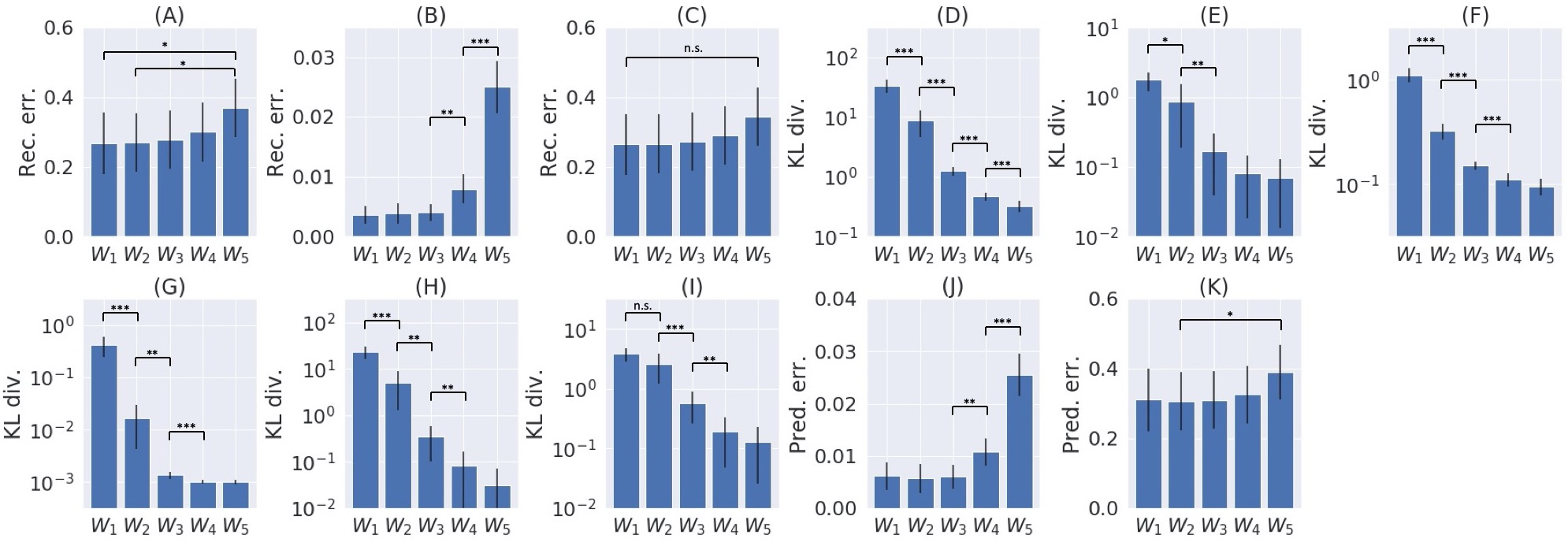}
    \caption{Reconstruction error, KL divergence minimization, and one-step, look-ahead prediction error in error-regression with five meta-prior settings. (A) Sum of reconstruction errors in proprioception and vision. (B) The reconstruction error in proprioception. (C) The reconstruction error in vision. (D) Sum of the KL divergence in all layers. (E) The KL divergence in the associative module. (F) The KL divergence in the slow layer of the proprioception module. (G) The KL divergence in the fast layer of the proprioception module. (H) The KL divergence in the slow layer of the vision module. (I) The KL divergence of the fast layer of the vision module. (J) One-step, look-ahead prediction error in proprioception. (K) One-step, look-ahead prediction error in vision. Error bars represent the standard deviation of 10 models with different weight initialization. Asterisks represent the statistical significance in t-tests: $*$ for $p<0.05$, $**$ for $p<0.01$, and $***$ for $p<0.001$. Note that each graph has a different scale.}
    \label{exp2-1}
\end{figure}

Results are summarized in Figure \ref{exp2-1}. As a whole, with smaller values of the meta-prior, the reconstruction error was minimized more (Figure \ref{exp2-1}(A)), and the KL divergence remained large (Figure \ref{exp2-1}(D)), whereas with larger values of the meta-prior, the KL divergence was minimized more (Figure \ref{exp2-1}(D)), and the reconstruction error remained large (Figure\ref{exp2-1}(A)). This tendency can also be seen in the local proprioception module and vision module, although the reconstruction error in the vision module was not significantly different. In the proprioception module, as values of the meta-prior increased, the reconstruction error in proprioception became large (Figure \ref{exp2-1}(B)), and the KL divergence became small, both in the slow layer (Figure \ref{exp2-1}(F)) and in the fast layer (Figure \ref{exp2-1}(G)). In the vision module, as values of the meta-prior increased, though the reconstruction error in vision did not increase as significantly (Figure \ref{exp2-1}(C), the KL divergence became small in both the slow layer (Figure \ref{exp2-1}(H)) and the fast layer (Figure \ref{exp2-1}(I)). The KL divergence in the associative module also increased as values of the meta-prior increased (Figure \ref{exp2-1}(E)). In addition, with smaller values of the meta-prior, the one-step, look-ahead prediction error was minimized in both proprioception (Figure \ref{exp2-1}(J)) and in vision (Figure \ref{exp2-1}(K)). 

This is because the KL divergence term in the evidence lower bound was weighted more for minimization than was the reconstruction error term.
In this situation, the posterior $q(\boldsymbol{z}_t|\boldsymbol{d}_{t-1},e_{t:T'})$ at each time step in the ER window approached its prior $p(\boldsymbol{z}_t|\boldsymbol{d}_{t-1})$ by modulating the adaptive value $\boldsymbol{a}_t$, which is fed into the computation of the posterior $q(\boldsymbol{z}|\boldsymbol{d}_{t-1},e_{t:T'})$, while the prior $p(\boldsymbol{z}_t|\boldsymbol{d}_{t-1})$ was less changed. 
This means that network dynamics were driven mainly by the prior, and were less affected by sensory inputs. Network dynamics become more egocentric by following the prior, which was less modified by looser regulation of the complexity term (i.e., more weighting for the KL divergence term). On the other hand, with tighter regulation (i.e., less weighting of the KL divergence term), network dynamics became more adaptive to changes or fluctuations of sensory inputs by freely modulating the posterior in the direction of error minimization without being much constrained by the prior. In this condition, the prior $p(\boldsymbol{z}_t|\boldsymbol{d}_{t-1})$ at each time step in the window also changes because the posterior $q(\boldsymbol{z}_{t-1}|\boldsymbol{d}_{t-2})$ at the previous time step, which is mapped to $p(\boldsymbol{z}_t|\boldsymbol{d}_{t-1})$ through $\boldsymbol{d}_t$ also changes.

\begin{figure}[ht]
    \centering
    \includegraphics[width=1.0\textwidth]{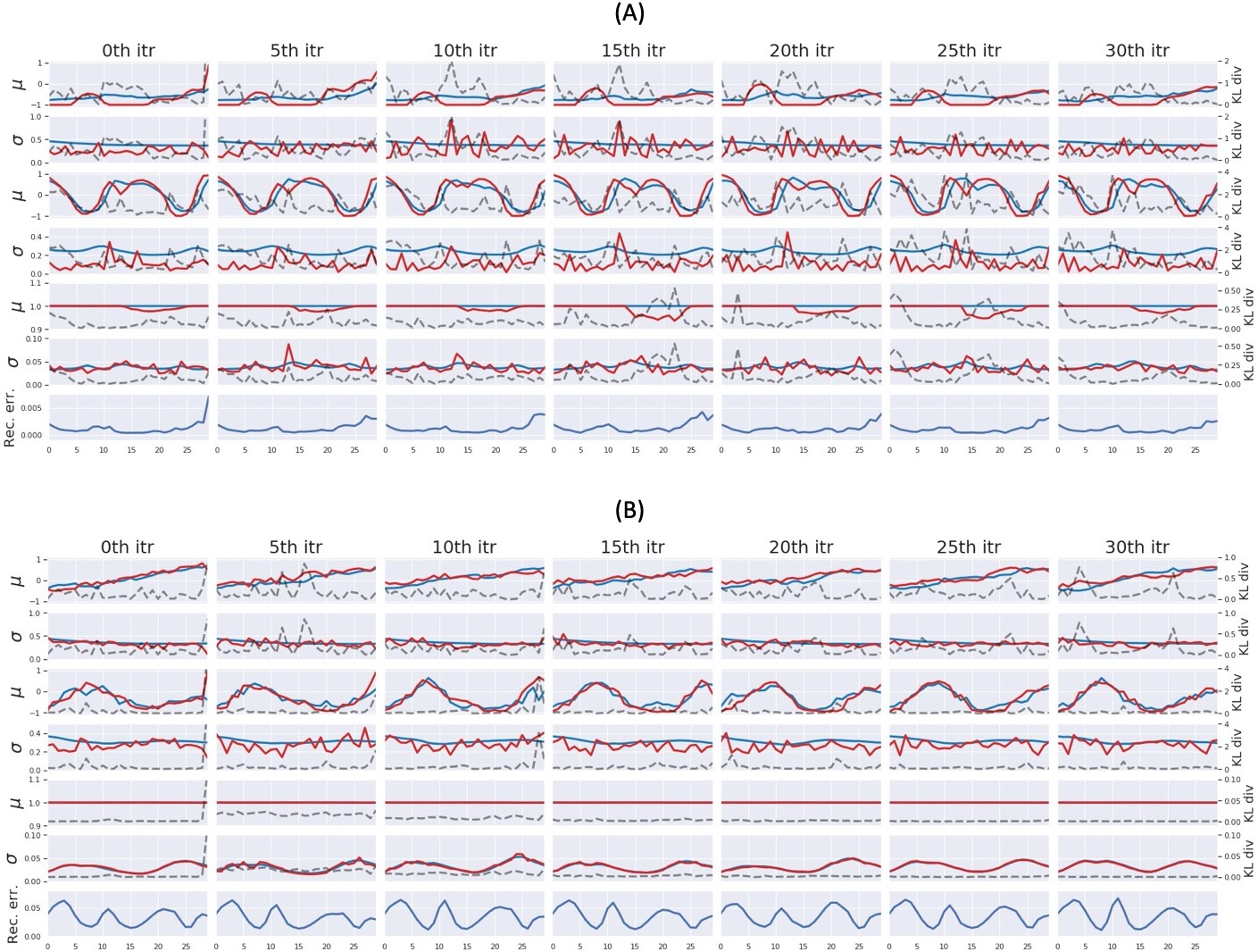}
    \caption{An example of the posterior inference during the pseudo imitative interaction in the $W_1$ setting (A) and in the $W_5$ setting (B). For clarity, only those parts involved in the proprioception module are shown. From the left, each column represents the network representation inside the ER window before the inference, after every 5th iteration up to the 30th iteration of the posterior inference. The first, third, and fifth rows show time trajectories of the mean $\boldsymbol{\mu}$ of the $\boldsymbol{z}$ in the associative module, the slow layer of the propriception module, and the fast layer of the proprioception module, respectively. The blue and red lines represent the prior $\boldsymbol{\mu}^p$ and the inferred posterior $\boldsymbol{\mu}^q$, respectively. The second, fourth, and sixth rows show the time trajectories of variance $\boldsymbol{\sigma}$ of $\boldsymbol{z}$ in the associative module, the slow layer of the proprioception module, and the fast layer of the proprioception module, respectively. The blue and red lines indicate the prior $\boldsymbol{\sigma}^p$ and the inferred posterior $\boldsymbol{\sigma}^q$, respectively. Dashed black lines indicate values of the KL divergence in each layer. The bottom row shows the reconstruction error at corresponding time steps.}
    \label{post_inf}
\end{figure}

\medskip
In the course of pseudo-imitative interaction, when the network observes a single time step of a new sensation, it infers sequences of the posterior inside the ER window with the aforementioned iterative computation of the error regression. 
Figure \ref{post_inf} displays some examples of the posterior inference during the process in which tight regulation of the complexity term ($W_1$ setting) (Figure \ref{post_inf} (A)) and loose regulation of the complexity term ($W_5$ setting) (Figure \ref{post_inf} (B)) are compared. For clarity, part of the network responsible for proprioception is shown. The columns illustrate, given a single time step of sensory observation, how the network inferred the posterior in terms of parameters of the posterior distribution, mean $\boldsymbol{\mu}$, and variance $\boldsymbol{\sigma}$ of multivariate Gaussian distributions under the effect of different values of the meta-prior through iterations. From the left, each column shows network dynamics before the inference, after 5th, 10th, 15th, 20th, 25th, and 30th iteration of the update of adaptive variable $\boldsymbol{a}$ inside the ER window with BPTT. The first, third, and fifth rows plot the relationship among the mean of the prior $\boldsymbol{\mu}^p$ (blue lines), the mean of the inferred posterior $\boldsymbol{\mu}^q$ (red lines), and the KL divergence (dashed black lines) in the associative module, in the slow and fast layers of the proprioception module, respectively. The second, fourth, and sixth rows plot the variance of the prior $\boldsymbol{\sigma}^p$ (blue lines), the variance of the inferred posterior $\boldsymbol{\sigma}^q$ (red lines), and the KL divergence (dashed black lines) in the associative module, in the slow and fast layers of the proprioception module, respectively. Although dimensions of $\boldsymbol{z}$ in the fast layer and in the slow layer of the proprioception module are greater than one, only one dimension is plotted for visibility.

In $W_1$ setting, the network is assigned smaller values of the meta-prior, which means that the complexity term is tightly regulated. Therefore, during the course of posterior inference, the inferred posterior is allowed to deviate somewhat from the prior to minimize the reconstruction error compared to the $W_5$ setting with looser regulation. This can be seen in Figure \ref{post_inf} (A). In the leftmost column, the network encountered a large reconstruction error in the last time step inside the ER window. This reconstruction error was eventually resolved while updating the posterior repeatedly as a result of distributing the KL divergence over the entire network in consideration of values of the meta-prior assigned to each layer. In the $W_1$ setting, the associative module had the smallest value of the meta-prior, the slow layer of the proprioception module had one with a moderate value, and the fast layer of the proprioception module had the largest value of the meta-prior. Thus, the largest discrepancy between the inferred posterior and the prior was allowed in the associative module and the smallest discrepancy in the fast layer of the proprioception module. This can be confirmed by comparing the posterior, the prior, and the value of KL divergence in each layer in Figure \ref{post_inf} (A).

In contrast, in the $W_5$ setting, the complexity term is loosely regulated with larger values of the meta-prior, which forces the network to keep the KL divergence small during the posterior inference. This can be observed in Figure \ref{post_inf} (B). During the iteration, the value of the KL divergence was strongly suppressed, and as a result, the reconstruction error remained large even after the posterior update. Compared to Figure \ref{post_inf} (A), the posterior was inferred so that it was closer to the prior (red lines are closer to blue lines).

\newpage

\begin{figure}[ht]
    \centering
    \includegraphics[width=1.0\textwidth]{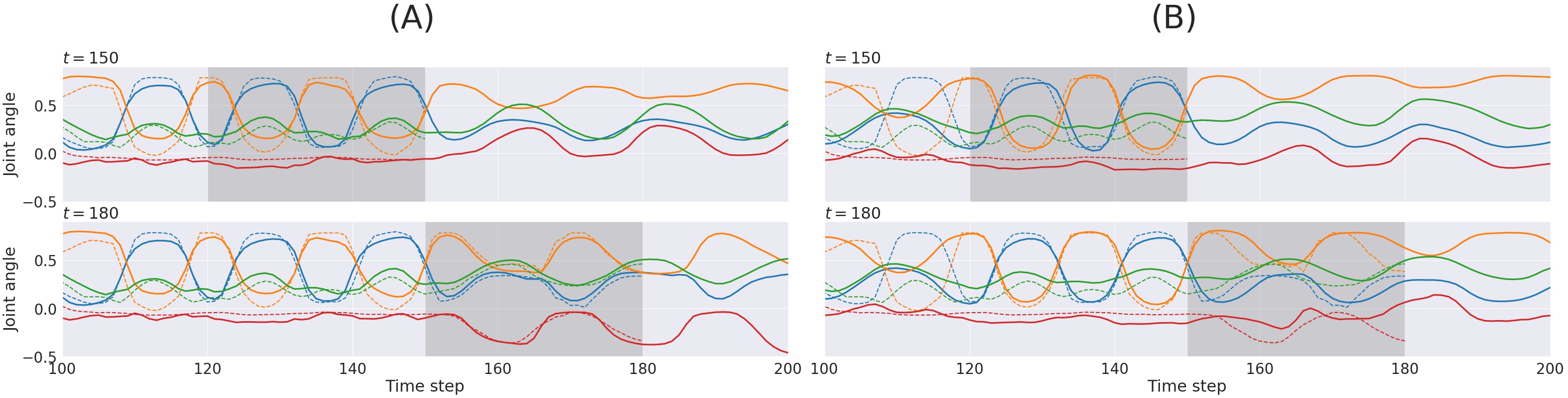}
    \caption{An example of time-series plots of neural activities in the output layer of the proprioception module in the $W_1$ setting (A) and in the $W_5$ setting (B). Reconstruction of the past observation and the future prediction at time step 150 (top) and at time step 180 (bottom) are shown. Solid lines represent prediction outputs, and dashed lines represent observations. The shadowed area indicates the error-regression window. For simplicity, only 4 of 16 joint-angles representing movements are shown.}
    \label{exp2-2}
\end{figure}

Figure \ref{exp2-2} (A) and (B) show examples of time-series plots of related neural activities of the proprioception module, comparing cases of tight ($W_1$ setting) and loose ($W_5$ setting) regulation of the complexity term. Both cases are computed for a situation observing the same visuo-proprioceptive sequence pattern. With tight regulation of the complexity term (Figure \ref{exp2-2} (A) top), the observation of the primitive A (dashed lines) was well reconstructed (solid lines) inside the ER window (gray area) from time steps 120 to 150, due to relatively stronger weighting of the accuracy term compared to the $W_5$ setting. Plots after time step 150 represent future predictions of the expectation of encountering the primitive B. From time steps 150 to 180 (Figure \ref{exp2-2} (A) bottom), the agent observed new sensory information where the primitive C instead of the predicted primitive of B was encountered. (Remember that there is a 50\% chance of encountering the primitive B or the primitive C.) This new observation was reconstructed inside the ER window. Based on the inferred posterior during this period, the robot updated the future prediction after time step 180 as the primitive C to be continued. Because of relatively stronger weighting in the accuracy term, the posterior was inferred to adapt to reality. The prediction was also updated accordingly (Figure \ref{exp2-2} (A) bottom). 

In the case of loose regulation (Figure \ref{exp2-2} (B) top), the observation was still well reconstructed inside the ER window. This is because primitive pattern A always follows either primitive pattern B or C so that it is easy to predict primitive A. Therefore, the reconstruction error inside the ER window was small from the beginning. Plots after time step 150 represent future predictions expecting primitive pattern B to be encountered. After observing new sensory information in which primitive pattern C instead of the predicted primitive pattern B was encountered between time steps 150 and 180 (Figure \ref{exp2-2} (B) bottom); however, the new observation was not reconstructed well inside the ER window. Due to tight regulation of the KL divergence term (loose regulation of the complexity term), the posterior was forced closer to the prior by ignoring the new observation. Consequently, the inferred posterior did not affect the prior as much as in the $W_1$ setting, which resulted in generation of consistent predictions for the future. Actually, the look-ahead prediction made at time step 150, shown in the top row, and the one made at time step 180 in the bottom row are almost the same. These observations imply that both the prediction of the future and the reflection of the past become more adaptive to sensory observation in the case of tighter regulation of the complexity term, whereas they become more persistent regardless of sensory observations in the case of looser regulation.

Some representative videos related to Experiment 2 can be seen at \href{https://youtu.be/WaODCuuVZjA}{video link A} and at  \href{https://youtu.be/rXDCOb7Q89I}{video link B} for the $W_1$ condition and the $W_5$ condition, respectively. These videos show how prediction of the future as well as reflection of the past can be performed for each condition. Also, further temporal details during the error-regression process can be seen at \href{https://youtu.be/I_pexMjyO4s}{video link C} and at  \href{https://youtu.be/0K7iQsE7ltU}{video link D} for the $W_1$ condition and the $W_5$ condition, respectively. In these videos, there is some divergence between the prior and the posterior in terms of mean and variance and they are dynamically changing inside the ER-W in the $W_1$ condition, whereas these two profiles approximate each other, showing relatively persistent patterns in the $W_5$ condition. These observations accord with our analysis, described previously.

\section{Discussion}
The current study investigated underlying mechanism of the strength of agency in social interaction by proposing a model for imitative interaction using multimodal sensation based on the framework of PC and AIF. We proposed a hypothesis that tightness used to regulate the complexity term in the evidence lower bound in the proposed model should contribute to the strength of agency. This hypothesis was evaluated by conducting simulation experiments on a pseudo-human-robot imitative interaction using the model.

First, we examined possible effects of changing the tightness used to regulate the complexity term for each sensory module during the learning phase in coordination and integration of different modalities of sensation and those in the test imitation phase. Our results showed that the complexity term in the vision module should be regulated more than that of the proprioception module. This is because vision and proprioception are significantly different with respect to their intrinsic randomness, as visual inputs fluctuate more due to optical conditions, such as illumination and surface reflectiveness. We concluded that the complexity term in the vision module should be regulated much more than that for the proprioception module to achieve better generalization in learning.

Next, we investigated the strength of agency as the main focus of the current study by changing the tightness used to regulate the complexity term for the entire network relative to that which was introduced in the training phase. For this purpose, characteristics of pseudo-imitative interaction were examined by scaling the meta-prior of each module equally to larger or smaller values using the network that had been evaluated as successful in the previous experiment. Our results demonstrated that changing the meta-prior this way affects performance characteristics of imitative interaction significantly. With looser regulation of the complexity term, the agent tends to act more egocentrically, without adapting to the other. In contrast, with tighter regulation of the complexity term, the agent tends to follow its human counterpart by adapting its internal state. This result implies that the strength of SoA can be modulated by adjusting the tightness with which the complexity term is regulated after the learning phase.

In the current study, we evaluated this hypothesis by considering a task of imitative interaction between a robot and a human counterpart. 
In such an imitative interaction, there could be two situations: the robot follows the human's movements, or the human follows the robot's movements. In our experimental results, the agent with tight regulation of the complexity term corresponded to the former case, and that with loose regulation to the latter. These findings could provide new insights into computational modeling studies of MNS. 
Our group's previous studies \citep{ahmadi2017can, hwang2020dealing} on modeling MNS using deterministic RNNs that were applied to robot imitation experiments, introduced a scheme similar to the ER scheme described in the current study, in the sense that both reinterpret past observations and update future predictions. In the model, deterministic latent variables at the onset time step of the immediate past window are updated by means of the ER scheme. Since these latent variables are not constrained by any prior probability distribution (unlike the sequential prior scheme), they adapt to sensory sequences encountered  for minimizing the error directly wherein the speed of updating is simply determined by the adaptation rate to update the latent variables.

On the other hand, in the case of the ER, which uses PV-RNN, the update of stochastic latent variables $\boldsymbol{z}$ at each time step inside the ER window are constrained by the sequential prior represented in terms of a Gaussian probability distribution. If the PV-RNN is developed more toward deterministic dynamics by setting the meta-prior with larger values, the sequential prior for each stochastic latent variable should have a peaky distribution with relatively small variance. In such a case, the approximate posterior cannot adapt to the sensory sequence by using the propagated error signal because the current prior is estimated with a strong belief. In contrast, if the PV-RNN is developed toward a more random process by setting the meta-prior with smaller values, at each time step the prior should exhibit a wide distribution with large variance. Then, the posterior can easily adapt to the sensation using the error signal, because the current prior is estimated with a weak belief. Therefore, the PV-RNN can show both mirror neuron-type adaptive response and egocentric behavior, depending on the setting of the meta-prior in interactions among agents. The deterministic RNN models shown in \cite{ahmadi2017can, hwang2020dealing}, however, can only show mirror neuron-like adaptive responses.

By following the above discussion, one essential advantage of using variational RNNs, such as PV-RNN, compared with conventional deterministic RNNs, is that they can predict not only future contents, but can also estimate predictability of such predictions or in other words, the credibility of prediction, as discussed in formulation of the {\it free-energy princple} \citep{friston2005theory}. This sort of cognitive competency of second order prediction by means of representing the belief of prediction, by which the strength of agency can be mechanized, provides modeling of agents, including cognitive robots with more complexity and richness in ways of interacting with other agents, as well as the physical world, as the current study demonstrates, at least partially.

In everyday social interactions, humans don't just follow others, nor do they lead them all the time. Rather humans sometimes follow others and sometimes lead them, depending on the moment-by-moment context or social situation. Psychological studies indicate that turn-taking between following and leading can occur quite spontaneously in various social cognitive behaviors, including conversation \citep{sacks1978simplest}, mother-infant pre-verbal communication \citep{trevarthen1979communication} and imitation \citep{nadel2002imitation}. In considering possible mechanisms underlying turn-taking, some researchers \citep{ikegami2007turn, ito2004line} suggest that turn-taking may develop due to potential instability, such as chaos formed in coupled dynamics between two agents in their modeling studies. We consider meta-level dynamics coupling two agents, whereby the value of the meta-prior to regulate the complexity terms in the two agents counteract one another mutually. This could result in autonomous shifts between the leading mode by increasing the meta-prior and the following mode by reducing it.

Future studies should examine the aforementioned mechanism for turn-taking by conducting an online experiment of human-robot interactions.
However, the computational cost of online error-regression for the posterior inference has been the major bottleneck for conducting such experiments in real time, and this is why the current study was limited to a simulation of pseudo-imitative interaction using recorded visuo-proprioceptive sequence patterns, rather than introducing actual, real-time, human-robot interaction. Although our group has shown that some real-time experiments using online ER are possible using only the sensory modality of proprioception \citep{chame2019cognitive}, it becomes prohibitive when also using vision, with sufficient pixel resolution. Regarding this problem, some may suggest employing other types of variational models, such as a variational recurrent neural network (VRNN) \citep{chung2015recurrent}, because a VRNN demands far less computation time, since the posterior at each time step can be inferred by simple sequential mapping of inputs using an autoencoder \citep{kingma2016improved}. However, the current scheme for inference of the posterior through iterative computation for optimization is probably vital for any embodied cognitive systems that require rapid adaptation of internal states to the environment. Actually, Ahmadi and Tani \citep{ahmadi2019novel} showed that PV-RNN performs better than VRNN in online prediction in dynamically changing environments by inferring the posterior using the error-regression scheme. Therefore, future studies should explore possible methods for accelerating online error-regression of the model, such as by massive parallelization so as to conduct real-time, human-robot interactions using the current model. 

\section*{Ethics statement}
Written informed consent was obtained from the individuals for publication of any potentially identifiable images or data included in this article.

\section*{Conflict of Interest Statement}
The authors declare that the research was conducted in the absence of any commercial or financial relationships that could be construed as a potential conflict of interest.

\section*{Author Contributions}
WO and JT conceived the concepts and models and contributed to the writing. WO conducted the experiments.

\section*{Funding}
This study was supported by funding from Okinawa Institute of Science and Technology Graduate University. This study has also been partially supported by a Grant-in-Aid for Scientific Research(A) in Japan, 20H00001, “Phenomenology of Altered Consciousness: An Interdisciplinary Approach through Philosophy, Mathematics, Neuroscience, and Robotics”.

\section*{Acknowledgments}
We thank lab members in the Cognitive Neurorobotics Research Unit. We are especially grateful to Ahmadreza Ahmadi and Prasanna Vijayaraghavan for their help in developing the model. We thank Siqing Hou for his help in collecting data. We also thank Steven D. Aird for editing the manuscript.

\bibliographystyle{frontiersinSCNS_ENG_HUMS}  
\bibliography{references}

\end{document}